\documentclass[journal]{IEEEtran}
\usepackage{latexsym}
\usepackage{graphicx}
\usepackage{amsfonts,amssymb,amsmath}
\usepackage{hyperref}
\def\BibTeX{{\rm B\kern-.05em{\sc i\kern-.025em b}\kern-.08em T\kern-.1667em\lower.7ex\hbox{E}\kern-.125emX}}

\newcommand{\eg}{e.\,g.,}
\newcommand{\ie}{i.\,e.,}

\newcommand{\RNum}[1]{\uppercase\expandafter{\romannumeral #1\relax}}

\usepackage{csquotes}
\usepackage{siunitx }
\usepackage{subfig}
\usepackage{booktabs}
\usepackage{threeparttable}
\usepackage{multirow}
\usepackage{hhline}
\usepackage{xcolor}

\begin{document}

\title{Interaction Detection Between Vehicles and Vulnerable Road Users: A Deep Generative Approach with Attention}

\author{Hao Cheng,~\IEEEmembership{Member, IEEE}, Li Feng, Hailong Liu,~\IEEEmembership{Member, IEEE}, Takatsugu Hirayama,~\IEEEmembership{Member, IEEE},\\ Hiroshi Murase,~\IEEEmembership{Fellow,~IEEE}, and Monika Sester
\thanks{Hao Cheng, Li Feng and Monika Sester are with Leibniz University Hannover, Appelstrasse 9a, 30167 Hannover Germany. (e-mail: cheng@ikg.uni-hannover.de; sester@ikg.uni-hannover.de and li.feng1@outlook.com).}
\thanks{Hailong Liu and Hiroshi Murase are with Graduate School of Informatics, Nagoya University, Furo-cho, Chikusa-ku, Nagoya, Aichi, 464-8601, JAPAN. (e-mail: lhl881210@live.com and murase@is.nagoya-u.ac.jp).}
\thanks{Takatsugu Hirayama is with University of Human Environments, 6-2, Kamisanbonmatsu, Motojuku-cho, Okazaki, Aichi, 444-3505, JAPAN . (e-mail: t-hirayama@uhe.ac.jp).}}

\maketitle

\begin{abstract}
Intersections where vehicles are permitted to turn and interact with vulnerable road users (VRUs) like pedestrians and cyclists are among some of the most challenging locations for automated and accurate recognition of road users' behavior. 
In this paper, we propose a deep conditional generative model for interaction detection at such locations.
It aims to automatically analyze massive video data about the continuity of road users' behavior. 
This task is essential for many intelligent transportation systems such as traffic safety control and self-driving cars that depend on the understanding of road users' locomotion.
A Conditional Variational Auto-Encoder based model with Gaussian latent variables is trained to encode road users' behavior and perform probabilistic and diverse predictions of interactions.  
The model takes as input the information of road users' type, position and motion automatically extracted by a deep learning object detector and optical flow from videos, and generates frame-wise probabilities that represent the dynamics of interactions between a turning vehicle and any VRUs involved. 
The model's efficacy was validated by testing on real--world datasets acquired from two different intersections. 
It achieved an F1-score above 0.96 at a right--turn intersection in Germany and 0.89 at a left--turn intersection in Japan, both with very busy traffic flows. 
\end{abstract}


\section{INTRODUCTION}
\label{sec:intro}
\begin{figure}[t!]
\centerline{\includegraphics[trim=0in 0.3in 0in 0.3in, clip=true, width=3.5in]{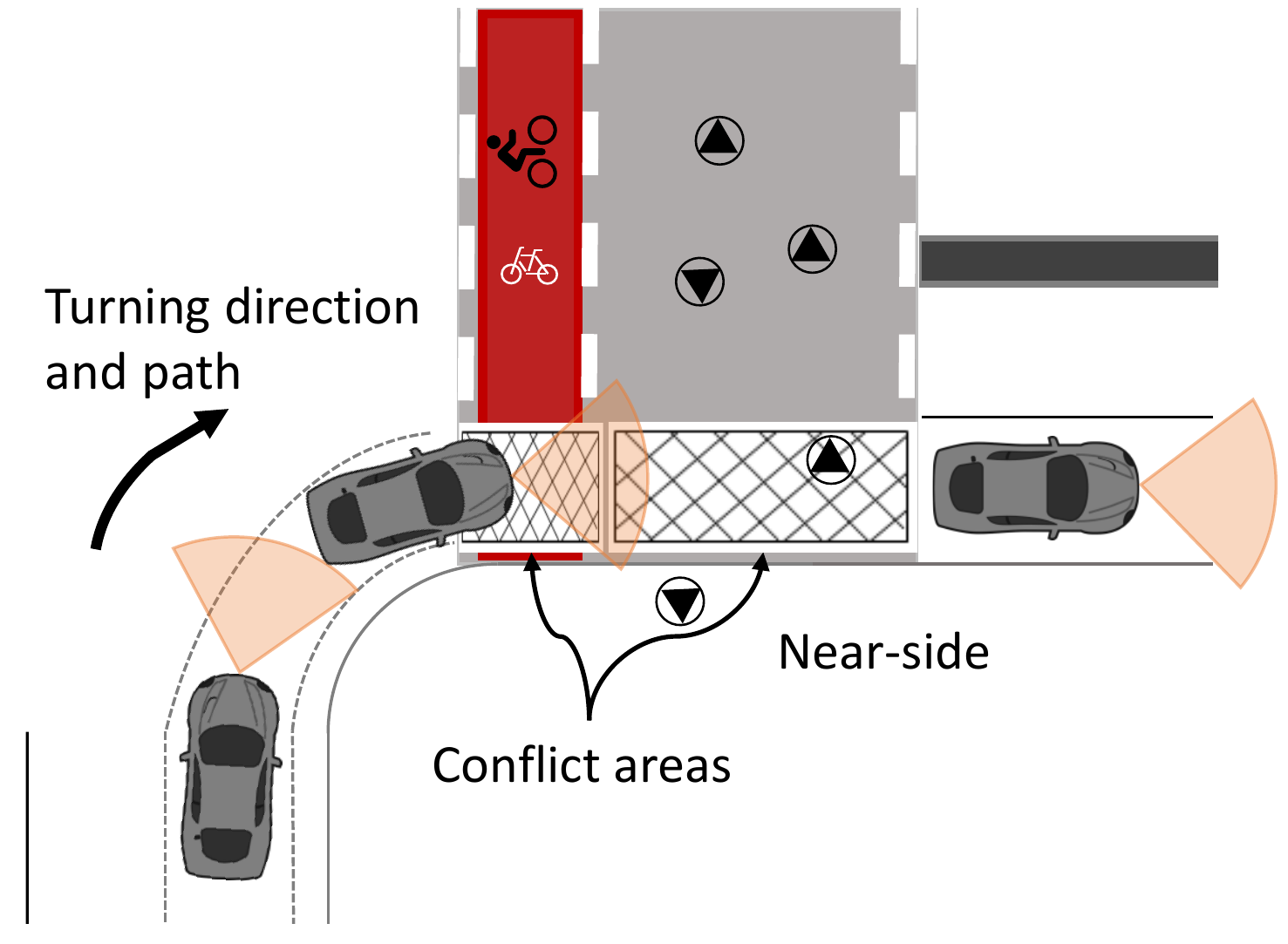}}
\caption{A right-turn intersection in Germany. A dedicated lane for cyclists is typically parallel to the crossing zone.}
\label{fig:intersection}
\end{figure}

\IEEEPARstart{I}n real--world traffic situations, it is not uncommon that heterogeneous road users like vehicles and vulnerable road users (VRUs, \eg~pedestrians and cyclists) have to directly interact with each other at some particular locations. 
Especially in city traffic, such type of locations include the turning areas of the so-called Turn-on-Red (TOR) intersections~\cite{mcgee1976right} or, more generally, intersections that allow vehicles to turn while other road users are crossing. 
During the time window of vehicles' turning, their behavior is largely guided by social protocols, \eg~right-of-way or courtesy. 
For example, in Germany, as shown in Fig.~\ref{fig:intersection}, a turning vehicle at a permissive right--turn intersection often encounters cyclists that are passing by and pedestrians that are cross walking in the conflict areas. In Japan, a similar situation can be found at a permissive left--turn intersection~\cite{alhajyaseen2012estimation} in left--hand traffic.

Efficiently and accurately learning how vehicles and VRUs interact with each other at such intersections is important for many applications.
As statistics show, accidents often occur at places where vehicles and VRUs confront each other and there have been reports that VRUs were seriously injured by the overlook of car drivers at turning intersections~\cite{choi2010crash,habibovic2011requirements,shirazi2016looking}. 
Thus, one important application is the analysis of interactions and critical situations. 
In addition, the foreseeable advent of autonomous driving in urban areas~\cite{franke1998autonomous}, particularly in such locations, requires accurate recognition of road users' behavior. Another potential application would be an accident warning system for road users.

Nowadays, with the ubiquity of traffic data and the development of computer vision techniques, there is a high chance for automatically recognizing road users' behavior from massive video data. 
Hence, in this paper we aim to investigate an efficient way that can automatically analyze whether the continuity of road users' behavior is interrupted when vehicles and VRUs meet at busy intersections, which we formalize as interaction detection by automatically extracting the user type, location and motion information from videos.

\begin{figure}[t!]
	\centering
	\includegraphics[trim=0in 0.3in 0.3in 0.0in, clip=true, width=2.5in]{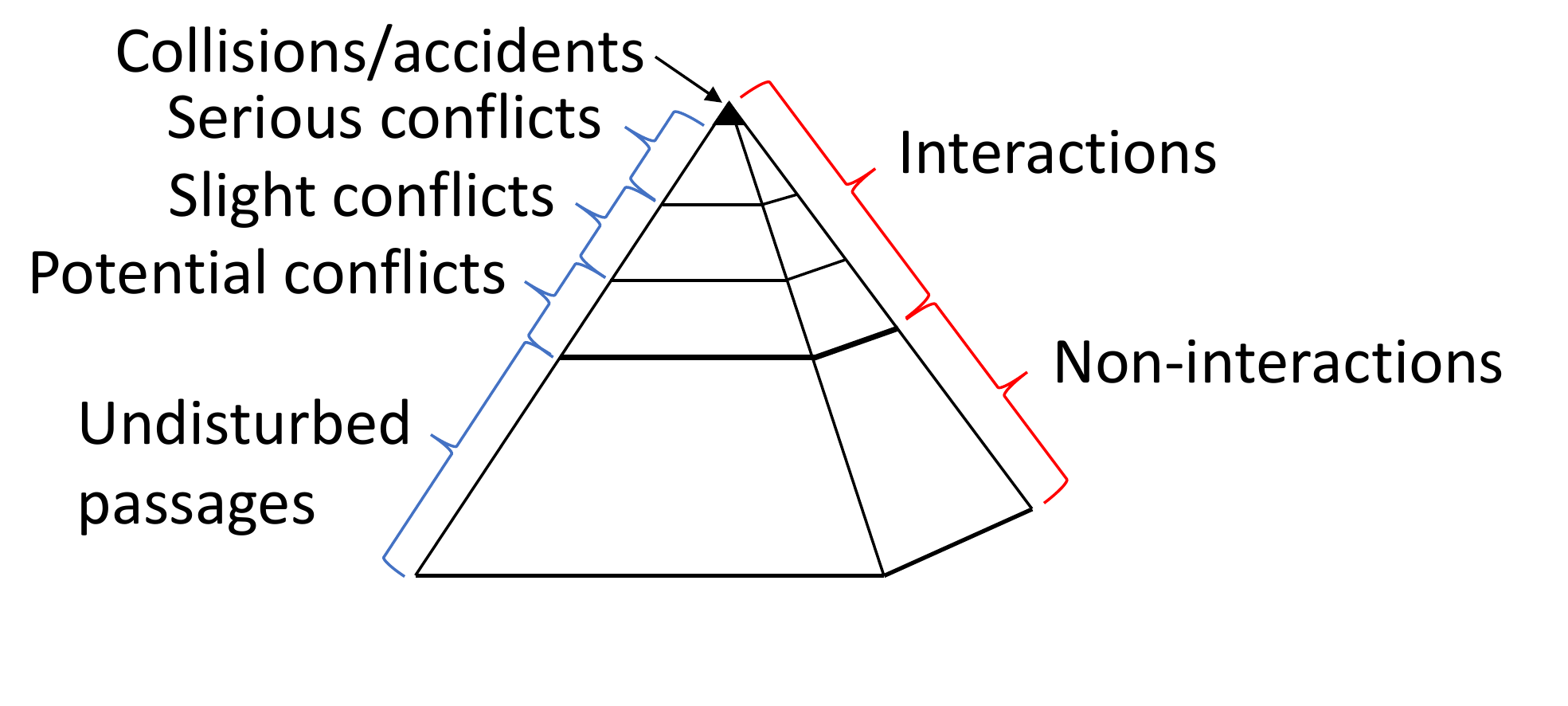}
	\caption[The pyramid of interactions]{The pyramid of interactions. The figure is partially adapted from~\cite{svensson2006estimating}.}
	\label{fig:rw-interactionpyramid}
\end{figure}

The concept of \textit{interaction} represents a changing level of reaction between road users. As defined by~\cite{saunier2008probabilistic} an interaction is ``a situation in which two or more road users are close enough in space and time and their distance is decreasing". Similarly, \cite{svensson2006estimating} describes an interaction between road users as ``a continuum of safety related events".
Moreover, \cite{svensson2006estimating, sayed1999traffic} relate interaction to conflict~\cite{perkins1968traffic} that interaction can range from collision to negligible conflict risks. However, in everyday traffic, collisions or accidents fortunately only account for a very small amount---the tip of the pyramid of interactions~\cite{svensson2006estimating}, see Fig.~\ref{fig:rw-interactionpyramid}. More frequent events are conflicts with different degrees of severity (serious, slight and potential) and undisturbed passages.  Therefore, in this paper, a high level of classification is adopted for classifying events into non-interactions (undisturbed passages) and interactions (all the other events). 

The task of \textit{interaction detection} is to differentiate interaction and non-interaction levels over the dynamics of a vehicle turning sequence. Interaction is needed if the turning vehicle drives into an intersection while any VRUs approaching or moving in the intersection space (see Fig.~\ref{fig:intersection}). In order to avoid any conflicts that might happen at any time during the vehicle's turning, they adapt their movement, \ie~velocity and orientation, accordingly. Otherwise, no interaction is needed if the target vehicle drives in an undisturbed manner with VRUs in its neighborhood, if there are any.

There are many challenges for automated interaction detection between vehicles and VRUs using video data.
Road users' behavior is dynamic and stochastic as they have to adjust their motion according to the reaction from each other. In addition, mixed types and varying numbers of roads users, as well as direct confrontations greatly complicate this task. The following open questions have to be addressed: (\RNum{1}) How to efficiently acquire, process, and label a large amount of video data for training a deep learning model for interaction detection considering all the relevant road users?
(\RNum{2}) How can a system automatically detect the location and motion of the involved road users?
(\RNum{3}) How to represent the dynamics of interactions in vehicle turning sequences of varying duration?

To tackle the above challenges, we propose a deep conditional generative model based on Conditional Variational Auto-Encoder (CVAE)~\cite{sohn2015learning} for automated interaction detection. The model is conditioned on the information extracted from video data and performs probabilistic inference for interaction prediction. As opposed to a disciminative model~\cite{cheng2020automatic} that distinguishes interaction classes (interaction vs. non-interaction) from the observed information, a set of Gaussian latent variables are used to encode the dynamic and stochastic behavior patterns, which enables the generative model to perform diverse predictions at inference time. The contributions of this work are summarized as follows:
\begin{itemize}
    \item[1)] Various activities among all road user types were recorded using a camera at a right--turn intersection in Germany and a left--turn intersection in Japan for very busy traffic flows. They were processed for interaction detection in both right-- and left--hand real--world traffic. In the future the data will be released for further research.
    \item[2)] We combine a deep learning object detector to automatically detect all the relevant road users and optical flow to extract their motion. The combination captures the dynamics of all the road users and circumvents the tremendous work of manual tracking of trajectories.
    \item[3)] Both sliding window and padding methods are explored to parse vehicle turning sequences of varying lengths. 
    \item[4)] We propose an end-to-end sequence-to-sequence conditional generative model with a self-attention mechanism~\cite{vaswani2017attention} for interaction detection, which simultaneously takes both the object and motion information sequences and generates probabilities of interaction at each short interval ({$<$\SI{0.1}{s}}). The probabilities change accordingly when the intensity of interaction changes between a turning vehicle and the involved VRUs over time.
\end{itemize}

The remainder of the paper is organized as follows. Sect.~\ref{sec:rw-interactiondetection} reviews the related studies on road user behavior at intersections. The proposed methodology is introduced in Section~\ref{sec:interactiondetection}. The detailed information of the datasets and evaluation metrics is provided in Sec.~\ref{sec:InteDetcExperimentSettings}. The experimental results are presented and analyzed in Sec.~\ref{sec:InteDetcResults} and the limitations of the model are further discussed in Sec.~\ref{sec:InteDetcDiscussion}. Finally, the conclusions are drawn in Sec.~\ref{sec:conclusion} with potential directions of future work.

\section{Related Work}
\label{sec:rw-interactiondetection}
Early studies on road user behavior at intersections are focused on collision and conflict analyses~\cite{allen1978analysis,compton1994safety,kaparias2010development,salamati2011development}. For example, \cite{allen1978analysis} manually observed and studied a total of 25 collision scenes taped by videos at an intersection over a period of one year. \cite{compton1994safety} conducted a study of the safety impact of permitting turns on red lights (TOR) based on data of crashes. Examining actual collision and conflict scenes has many limitations.  First, the occurrence of crash and accident events are very rare in daily traffic and vary from case to case, which cannot represent the majority of road users' behavior~\cite{sayed2013automated} as undisturbed passages are not included. Second, as pointed out by~\cite{ismail2009automated}, collision--based safety analysis is a reactive approach, which requires a significant number of collisions to be collected before an action is warranted.
Third, those data are likely to be incompletely documented or protected for legal and privacy reasons, which leads to the data acquisition being complicated or even not possible. Most importantly, the above drawbacks make it almost impossible to automatically analyze the behavior of road users.

The development of computer vision techniques allows for automated analysis of road users' behavior at intersections. The work by Ismail et al.,~\cite{ismail2009automated}, one of the early studies using computer vision techniques, automatically analyzed pedestrian--vehicle conflicts at an intersection using the extracted trajecotries from video data. 
Later on, similarly, several works used trajectories extracted from videos to analyze before--and--after vehicle-pedestrian conflicts~\cite{ismail2010automated}, in street designs with elements of shared space~\cite{kaparias2013analysis}, in less organized traffic environments~\cite{tageldin2016developing}, and vehicle--bicycle conflicts at intersections~\cite{sayed2013automated}. 
The work carried out by Ni et al.,~\cite{ni2016evaluation} analyzed pedestrian--vehicle interaction patterns using indicators of Time-to-collision (TTC)~\cite{hayward1972near} and Gap Time (GT)~\cite{ismail2009automated}.
First, trajectories were extracted by the semi-automated image processing tool Traffic Analyzer~\cite{suzuki2006trafficanalyzer}, and then according to the TTC and GT values derived from the trajectory speed profiles interactions were classified into three classes: hard interaction, soft interaction and no interaction. On the one hand, their work is very close to the studies carried out in this paper. Both interactions and non-interactions are studied at permissive right--turn intersections. On the other hand, their work is not fully automated in terms of trajectory extraction. Acquiring reliable trajectory data is often costly and time--consuming. The quality of the data is difficult to guarantee. For example, tracking multiple objects from frame to frame is very challenging due to, \eg~abrupt object motion, change of appearance and occlusions \cite{yilmaz2006object}. Errors and failures in detection will propagate to the process of tracking, which later will directly lead to wrong conclusions in the analysis step~\cite{sayed2013automated}. 
Moreover, the above works only consider either vehicle-pedestrian or vehicle-cyclist interactions. In real--world traffic situations at big intersections other heterogeneous road users are often involved at the same time.

In very recent years, deep learning methods have been successfully applied to understand road users' behavior at intersections using video data, whereas many of them~\cite{rasouli2017they,ghori2018learning,hoy2018learning,rasouli2019autonomous} are conducted from a perspective of a self-driving car for pedestrian intent detection. 
With respect to a third--person perspective, \cite{cheng2020automatic} trained an encoder-decoder model to automatically detect interactions using sequences of video frames from a static camera facing a very busy left--turn intersection in a Japanese city. However, the discriminative model is trained to optimize the reconstruction loss between the pairs of ground truth and prediction, which tends to learn the ``average'' behavior of road users. The dynamics and stochastic behavior patterns are not fully captured. Hence, in this paper, we propose a CVAE--based model with Gaussian latent variables to encode various behavior patterns and perform diverse predictions. We not only test our model in left--hand traffic but also right--hand traffic in different countries for interaction detection between vehicles and all the other VRUs. 

\section{Methodology}
\label{sec:interactiondetection}
This section explains the methodology of interaction detection in detail. 
Sec.~\ref{subsec:interactionproblemformulation} formulates the problem, 
Sec.~\ref{subsec:featureextraction} describes the extraction of the input features, 
Sec.~\ref{subsec:cvaeclassifier} introduces the detection model, and
Sec.~\ref{subsec:interactionuncertainty} provides the estimation of the model's uncertainty.

\begin{figure}[hbpt!]
	\centering
	\includegraphics[trim=0.6in 0.1in 0.7in 0.1in, clip=true, width=3.5in]{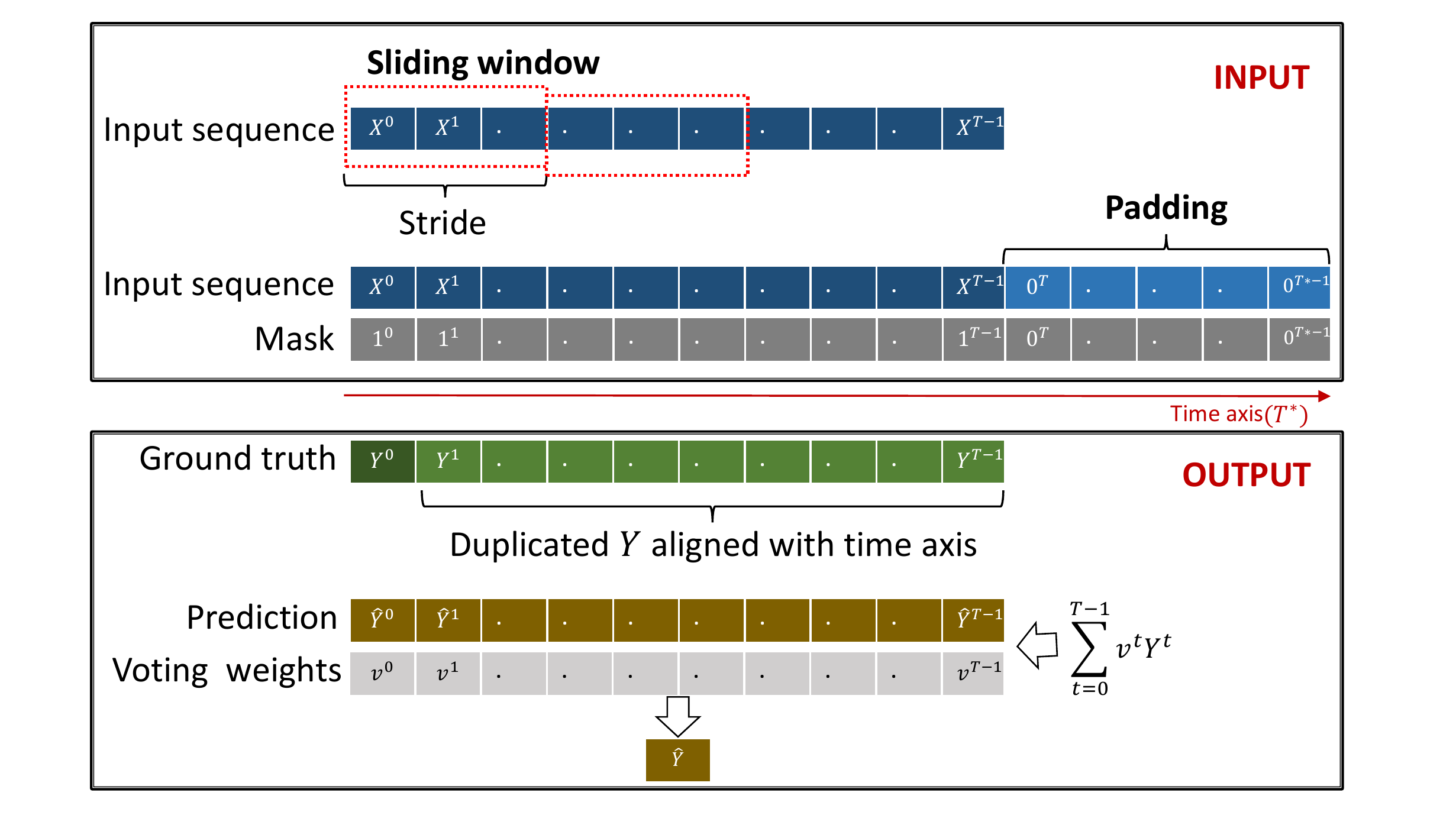}
	\caption[Sequence-to-sequence modeling using sliding window or padding method]{Sequence-to-sequence modeling using sliding window or padding method.}
	\label{fig:seqtoseq}
\end{figure}

\begin{figure*}[hbpt!]
	\centering
	\includegraphics[trim=0in 1in 0in 0in, width=\textwidth]{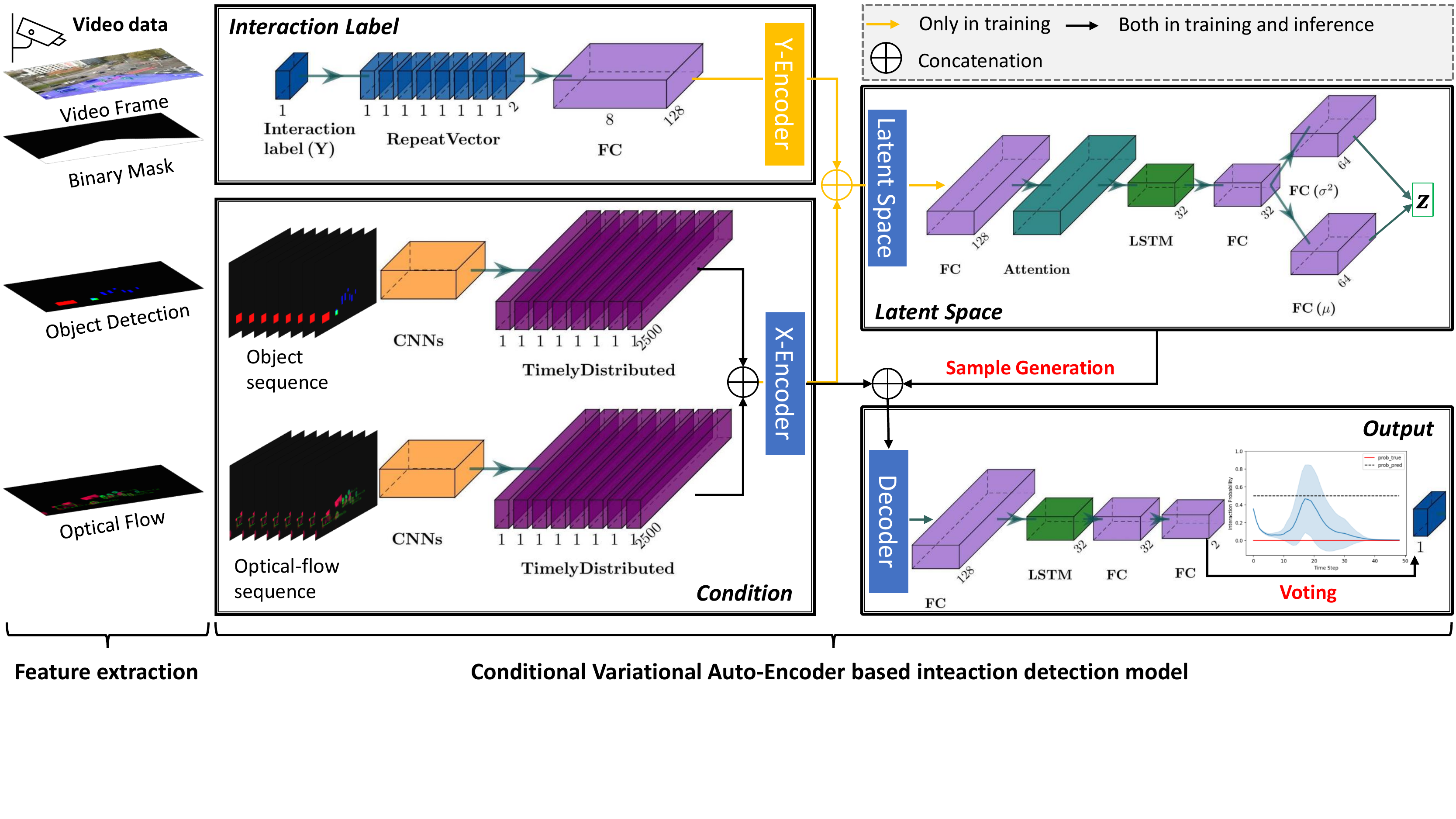}
	\caption[The pipeline of interaction detection]{The pipeline for interaction detection.}
	\label{fig:pipelineInteractionDetection}
\end{figure*}

\subsection{Problem formulation}
\label{subsec:interactionproblemformulation}

Interaction detection is formulated as a classification problem using the information extracted from videos. Given a set of observed vehicle turning sequences of video frames $\mathbf{X}=\{\boldsymbol{X}_1, \cdots,\,\boldsymbol{X}_i,\cdots\}$, the \textit{input} of the $i$-th sequence is characterized as $\boldsymbol{X}_i^{(T)} = \{X_i^t\}_{t=0}^{T-1}$, where $X^t \in \mathbb{R}^{W{\times}H{\times}C}$ is the frame at time step $t$, and $T$ is the total number of observed frames for the sequence. $W$, $H$ and $C$ denote the width, height and the number of channels of each frame. Instead of using raw images, object and optical--flow information (see Sec.~\ref{subsec:featureextraction} for more details) is extracted from the frames and is used as the sequence of the input. 
In this way, the personal information of road users,\eg face, gender, age, driving plates can be protected.
$\boldsymbol{Y}_i$ is the corresponding \textit{ground truth} interaction label and $\hat{\boldsymbol{Y}}_i$ is the \textit{prediction}. 

Moreover, sequence-to-sequence modeling is applied to learn the frame--wise dynamics of interactions over a turning sequence.
Similar to~\cite{ghori2018learning}, the task defined above is a weakly supervised learning problem due to the labelled data structure; The interaction label is a dichotomous class that represents the interaction level of the whole sequence. It does not provide detailed information about how the interaction level changes with time. In fact, it is not feasible to manually label each frame due to the tremendous amount of work. Without knowing the exact fine--grained frame--wise interaction label, the sequence--wise label is duplicated at each frame.
Hence, the form of the output is converted to obtain the time steps, denoted as $\boldsymbol{Y}_i^{(T)} = \{Y_i^t\}_{t=0}^{T-1}$ for the $i$-th turning sequence. 
Thereafter, the input and output are aligned at each frame for a sequence-to-sequence modeling, denoted by Fig.~\ref{fig:seqtoseq} and Eq.~\eqref{eq.seqtoseq}. 
\begin{equation}
\label{eq.seqtoseq}
    f(\boldsymbol{Y}_i|\boldsymbol{X}_i) = \lambda\sum_{t=0}^{T-1}f(\boldsymbol{Y}_i^{({T})}|\boldsymbol{X}_i^{({T})}),
\end{equation}
where $f$ denotes the detection model and $\lambda$ is a voting scheme that summaries the frame--wise predictions to the sequence--wise prediction. In this paper an average voting scheme that weighs the prediction at each frame equally is adopted and the sequence--wise prediction is the class label voted by the majority~\cite{cheng2020automatic}.
The above conversion is based on the following hypotheses: 
(1) Over a large dataset, sequence lengths vary from one to another, which provides rich interaction information of both long and short sequences. 
(2) The prediction error between $\boldsymbol{Y}_i$ and $\hat{\boldsymbol{Y}}_i$ is still computed at the sequence level because each frame--wise prediction only partially contributes to the sequence--wise prediction using the voting scheme. This mechanism enables the model to automatically learn the frame--wise dynamics at each frame in training time. 

In addition, sliding window and padding methods are proposed to deal with varying sequence lengths for training a recurrent neural network (RNN) based CVAE model, as shown in Fig.~\ref{fig:seqtoseq}. This is because the most commonly used RNNs, \eg~Long Short-Term Memory (LSTM)~\cite{hochreiter1997long}, for sequence modeling often require a fixed sequence length. However, at an intersection some vehicles can quickly complete the turning if the space happens to be free, whereas some vehicles may have to wait for a long time to let VRUs cross first.

The \textit{sliding window} method parses each sequence with a fixed window size $\mathsf{w}$. Eq.~\eqref{eq:slidingwindow} denotes the sliding window method with a stride being the same as $\mathsf{w}$. The overlap between two consecutive windows is allowed when the stride is set to be smaller than the window size.
\begin{equation}
\label{eq:slidingwindow}
    \boldsymbol{X_i}^{(T)} = \{X_i^{0},\cdots,\, X_i^{\mathsf{w}1}, \cdots,\, ~X_i^{\mathsf{w}k}\}, \text{~where~} k = \frac{T}{\mathsf{w}}.
\end{equation}

The \textit{padding method} uses zero-paddings at the end to extend the sequences shorter than a predefined length $T^\ast > T$. The value of $T^\ast$ can be adjusted to cover most of the sequences, \eg~$T^\ast = \text{Max}\{{T}_1, \cdots,~{T}_i,~\cdots\}$, where ${T}_i$ denotes the length of an arbitrary vehicle turning sequence. Meanwhile, a padding mask is used to annotate the exact sequence length so that the padded zero values are treated differently to mitigate the negative impact on the learning process. 
\begin{align}
\begin{split}
    \boldsymbol{X}_i^{(T)} &= \{X_i^0, \cdots, ~X_i^{T-1}, ~0^{T}~, \cdots, \,0^{T^\ast-1}\}, \\
    \text{Mask}_i^{(T)} &= \{1_i^0, \,\cdots, ~1_i^{T-1}, ~0^{T}~, \cdots, \,0^{T^\ast-1}\}. \\
\end{split}
\end{align}

After the formulation of the above sequence-to-sequence problem, we now introduce the pipeline of the proposed method for interaction detection, denoted by Fig.~\ref{fig:pipelineInteractionDetection}. It consists of two components: \textit{feature extraction} (Sec.~\ref{subsec:featureextraction}) and the sequence-to-sequence CVAE model (Sec.~\ref{subsec:cvaeclassifier}). Each component is explained in detail in the following subsections. 

\subsection{Feature extraction}
\label{subsec:featureextraction}
\begin{figure}[hbpt!]
\centering
    \subfloat[Object detection]{
    	\label{subfig:detection}
    	\hspace{-0.1cm}\begin{minipage}{1.7in}
    		\centering
    		\includegraphics[trim=0in 0in 0in 0.5in, clip=true, width=\textwidth]{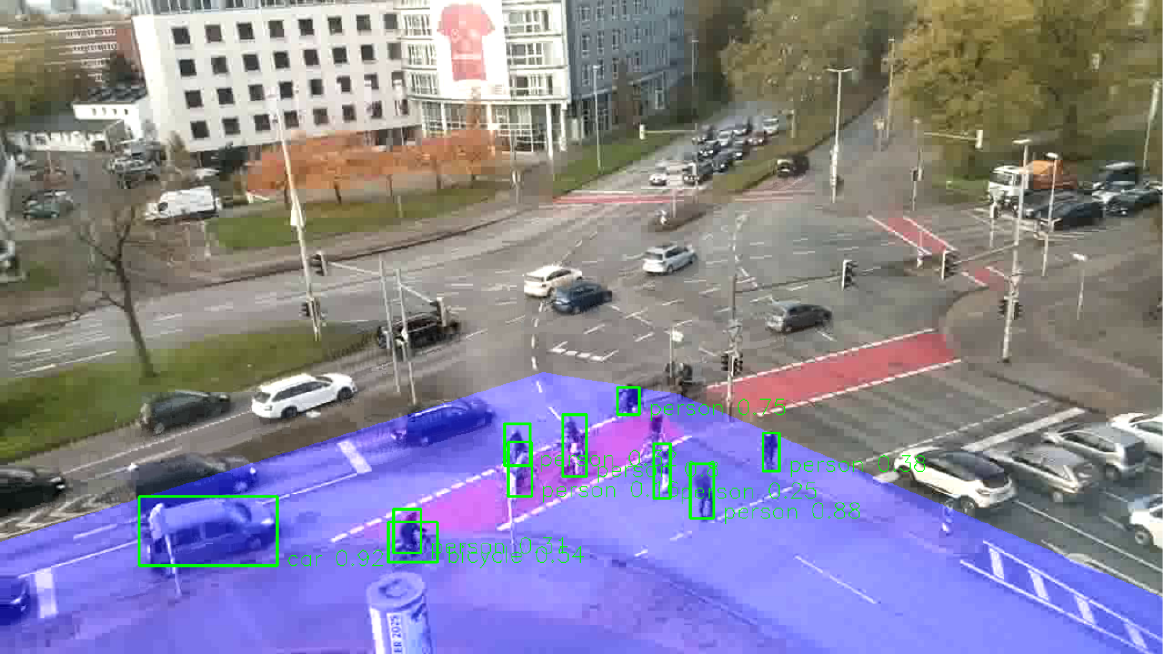}
    	\end{minipage}
    }%
    \subfloat[Binary mask]{
    	\label{subfig:mask}
    	\hspace{-0.1cm}\begin{minipage}{1.7in}
    		\centering
    		\includegraphics[trim=0in 0in 0in 1.45in, clip=true, width=\textwidth]{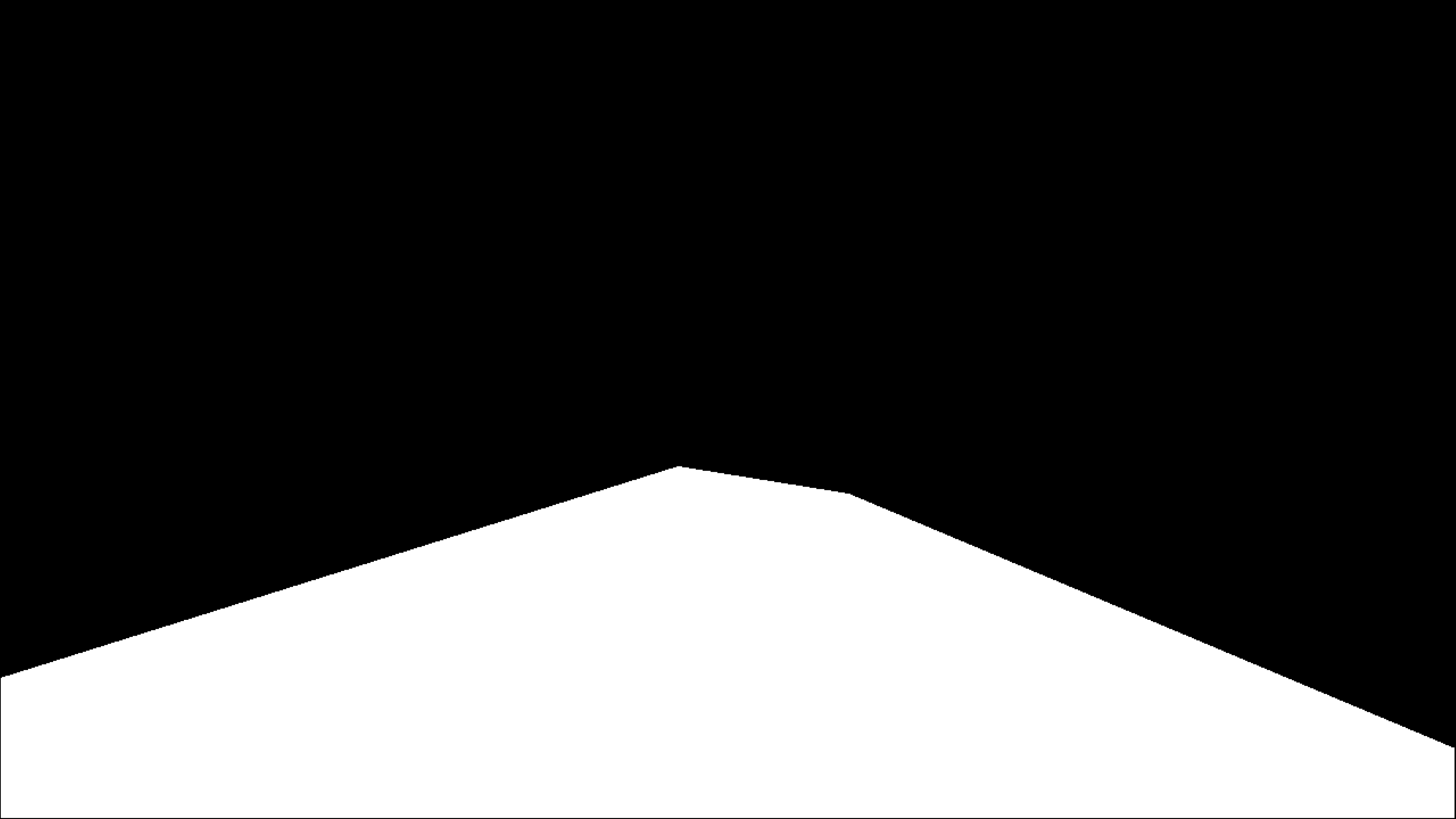}
    	\end{minipage}
    }%
    
    \subfloat[Object information]{
    	\label{subfig:bbox}
    	\hspace{-0.1cm}\begin{minipage}{1.7in}
    		\centering
    		\includegraphics[trim=0in 0in 0in 0.5in, clip=true, width=\textwidth]{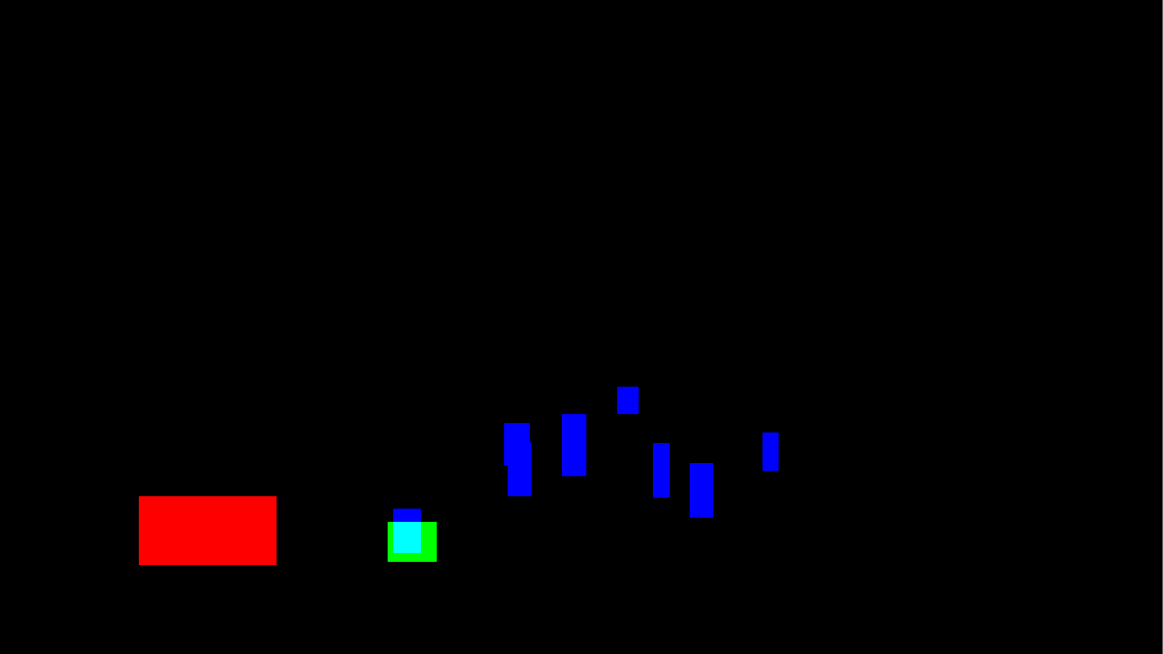}
    	\end{minipage}
    }%
     \subfloat[Optical--flow information]{
    	\label{subfig:optical_flow}
    	\hspace{-0.1cm}\begin{minipage}{1.7in}
    		\centering
    		\includegraphics[trim=0in 0in 0in 0.5in, clip=true, width=\textwidth]{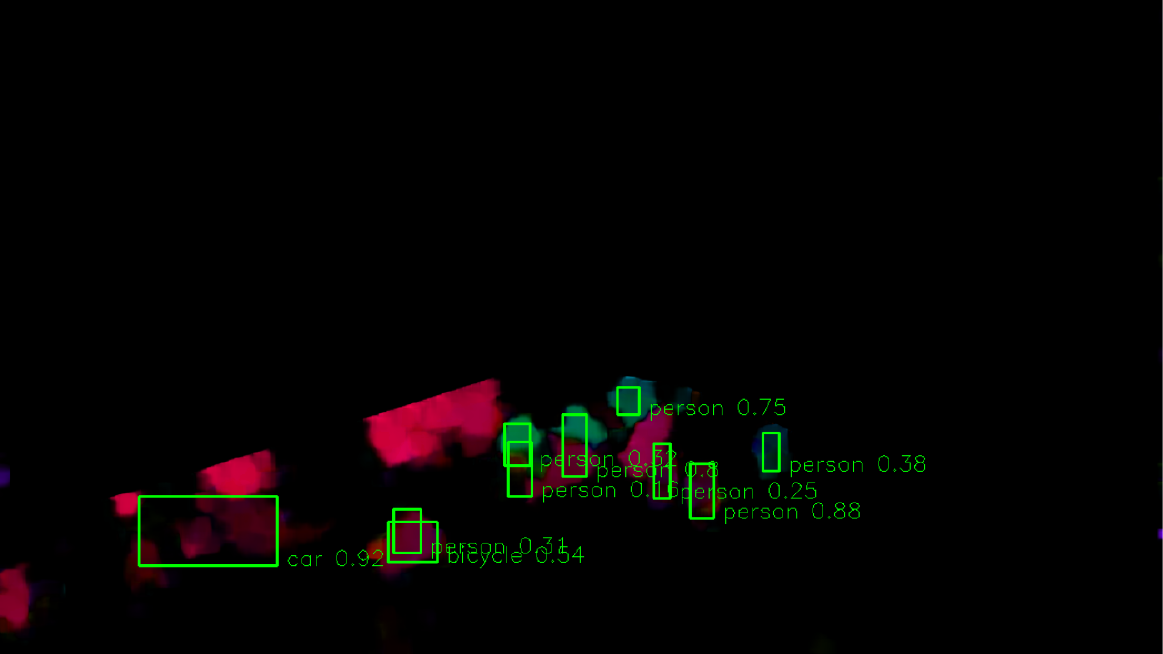}
    	\end{minipage}
    }
\caption[Input features for interaction detection]{Input features for interaction detection. Note that (c) only exemplifies three channels with pedestrians denoted in blue, bicycle(s) in green and car(s) in red color. The overlaid bounding boxes in (d) only serve the purpose of showing the location of the objects, including the static ones. They are not integrated into the optical--flow information.}  
\label{fig:inputfeaturesfordetection}
\end{figure}

Object information and optical--flow information extracted from video frames are used as input features for the interaction detection task, as shown in Table~\ref{tb:extractedfeatures} and Fig.~\ref{fig:inputfeaturesfordetection}.

\begin{table}[hbpt!]
\caption[Object and optical--flow information]{Object and optical flow information extracted by an object detector and the dense optical flow, respectively.}
\centering
\setlength{\tabcolsep}{4pt}
\begin{tabular}{llllll}
\toprule
Feature             & C1   & C2           & C3   & C4 & value      \\ \hhline{======}
Object             & pedestrians & bikes/motors & cars/trucks & buses     & \{0, 1\}        \\
Optical flow$^{*}$          & orientation & 1                    & velocity    &           & {\,[}0, 1{\,]} \\ \bottomrule
\end{tabular}
\begin{tabular}{@{}c@{}}
\multicolumn{1}{p{3.4in}}{$^{*}$The HSV (Hue, Saturation, Value) color representation is used to store the optical--flow information. The hue channel (C1) is used to store orientation, the saturation channel (C2) is set to its maximum, and the value channel (C3) is used to store velocity. Note that there are four channels in each object frame and only three channels in each optical--flow frame.} \\
\end{tabular}
\label{tb:extractedfeatures}
\end{table}

Object information contains road users' type and location. The deep learning object detector, such as YOLOv3 \cite{redmon2016you} or M2Det \cite{zhao2019m2det}, is leveraged for detecting all the relevant road users at each frame. Namely they are pedestrians, cyclists, motorbikes, cars, trucks and buses. Different channels are used to store the road--user position information and each channel is dedicated to one or two similar road user types; Based on the acquired data that only very few motorbikes were detected, they are stored in the same channel for bicycles. Cars/trucks are stored in one channel given their very similar trace of turning, see Table~\ref{tb:extractedfeatures}. The location of the detected road users (Fig.~\ref{subfig:detection}) is mapped  by the corresponding bounding boxes with values of one in each frame (Fig.~\ref{subfig:bbox}). Areas with no detected objects are set with values of zero, as shown in black color.

Optical flow is used to capture the motion of road users. It describes the distribution of velocities of moving pixels' brightness in two consecutive images \cite{horn1981determining}. Moving objects are captured by optical flow and static objects and background are ignored. The dense optical--flow algorithm~\cite{farneback2003two} is applied to map the displacement of moving objects and remove the static background information, see Fig~\ref{subfig:optical_flow}. Similarly, respective frame channels are dedicated to the information of orientation and velocity of the moving objects, see Table~\ref{tb:extractedfeatures}.

The area of interest is the tuning space of the intersection and is marked by a binary mask (Fig.~\ref{subfig:mask}), the other areas are not considered.
As shown in Fig.~\ref{subfig:detection} and \ref{subfig:mask}, the mask of the area of interest slightly extends into the through lane next to the turning lane.  Due to the oblique view of the camera, the upper body of the vehicles in the turning lane are partially projected into the through lane. The extended mask aims to include the upper body of the turning vehicles to be detected as well. The lower middle point of the bounding boxes of the detected vehicles is used to filter out the vehicles in the through lane. 
However, the extended mask introduces noise to the optical--flow information. For instance, as shown in Fig.~\ref{subfig:optical_flow}, the motion of the vehicles in the through lane is also captured by the optical--flow, which cannot be easily filtered out given the irregular shapes and occlusion. But it later turns out that this noise is not problematic when both the object information and the optical--flow information are combined as the input for training the interaction detection model. Interactions between vehicles and VRUs only happen in the crossing zone. 

\subsection{CVAE model for interaction detection}
\label{subsec:cvaeclassifier}
The model of predicting the probabilities of interaction between a turning vehicle and the other crossing road users is denoted as $f(\boldsymbol{Y}|\boldsymbol{X})= \text{arg\,max}_{\boldsymbol{Y}}~p(\boldsymbol{Y}|\boldsymbol{X},\, \mathbf{z})$, where $f$ is a CVAE model that performs probabilistic prediction and $\mathbf{z}$ are the Gaussian latent variables. The model encodes the information of interaction into a latent space and predicts the interaction label $\hat{\boldsymbol{Y}}$ conditioned on the input $\boldsymbol{X}$ and $\mathbf{z}$. The variational lower bound~\cite{sohn2015learning} of the model is given as follows:
\begin{align}
\begin{split}
    \log p_\theta(\boldsymbol{Y}|\boldsymbol{X}) \geq & -D_{KL}(q_\phi(\mathbf{z}|\boldsymbol{X}, \,\boldsymbol{Y})||p_\theta(\mathbf{z})) \\
                         & + \mathbb{E}_{q_\phi(\mathbf{z}|\boldsymbol{X}, \,\boldsymbol{Y})}
                          [\log p_\theta(\boldsymbol{Y}|\boldsymbol{X}, \,\mathbf{z})]. \label{eq:cvaeclassifier}
\end{split}
\end{align}
The model jointly trains a recognition model $q_\phi(\,\cdot\,)$ (a.\,k.\,a. encoder), and a generative model $p_\theta(\,\cdot\,)$ (a.\,k.\,a. decoder). In the training phase, the model is optimized via stochastic backpropogation~\cite{rezende2014stochastic}.  $q_\phi(\,\cdot\,)$ encodes the observed information and the ground truth label into the latent variables $\mathbf{z}$. In other words, the inserted label in training is combined with the condition to parameterize the Gaussian latent space, which later can be used for structured prediction to map the many possible outputs~\cite{sohn2015learning}.
$p_\theta(\,\cdot\,)$ decodes the prediction of the interaction label conditioned on the input and the latent variables. 
$-D_{KL}(\,\cdot\,)$ is the negative Kullback-Leibler divergence of the approximate posterior from the prior $p_\theta(\mathbf{z})$ and acts as a regularizer, which pushes the approximate posterior $q_\phi(\,\cdot\,)$ to the prior distribution $p_\theta(\mathbf{z})$. Note that in our model the prior is relaxed to make the latent variables statistically independent from the input variables so that $p_\theta(\mathbf{z}) = p_\theta(\mathbf{z}|\boldsymbol{X})$~\cite{DBLP:conf/nips/KingmaMRW14}.  The prediction loss $\mathbb{E}_{q_\phi(\mathbf{z}|\boldsymbol{X}, \,\boldsymbol{Y})}(\,\cdot\,)$ measures the distance between $\hat{\boldsymbol{Y}}$ and $\boldsymbol{Y}$. The binary cross--entropy loss is used as the prediction loss, as denoted by Eq.~\eqref{eq:binarycrossentropy}. 
\begin{equation}
\label{eq:binarycrossentropy}
    \mathcal{H}(\hat{\boldsymbol{Y}}-\boldsymbol{Y}) = -\{\boldsymbol{Y} \log\hat{\boldsymbol{Y}}
    + (1-\boldsymbol{Y}) \log(1-\hat{\boldsymbol{Y}})\}.
\end{equation}

In the inference phase, the decoder predicts the interaction label conditioned on the input of the observed information concatenated with a latent variable directly sampled from the Gaussian prior $p_\theta(\mathbf{z})$. The sampling process is done multiple times to perform diverse predictions~\cite{sohn2015learning}.

Convolutional Neural Networks (CNNs) and RNNs, as well as the self-attention mechanism~\cite{vaswani2017attention} are employed in the CVAE model to learn the parameters of $\theta$ and $\phi$.
As shown in Fig.~\ref{fig:pipelineInteractionDetection}, the encoder has two branches: X-Encoder and Y-Encoder. They are dedicated to extracting low--level features from the condition (the object and optical--flow information) and the interaction label information, respectively. Each module, \ie~X-Encoder, Y-Encoder, Latent Space, and Decoder of the CVAE model, is explained in detail as follows.

\begin{figure}[t!]
	\centering
	\includegraphics[trim=0in 0.5in 0in 0in, width=3.5in]{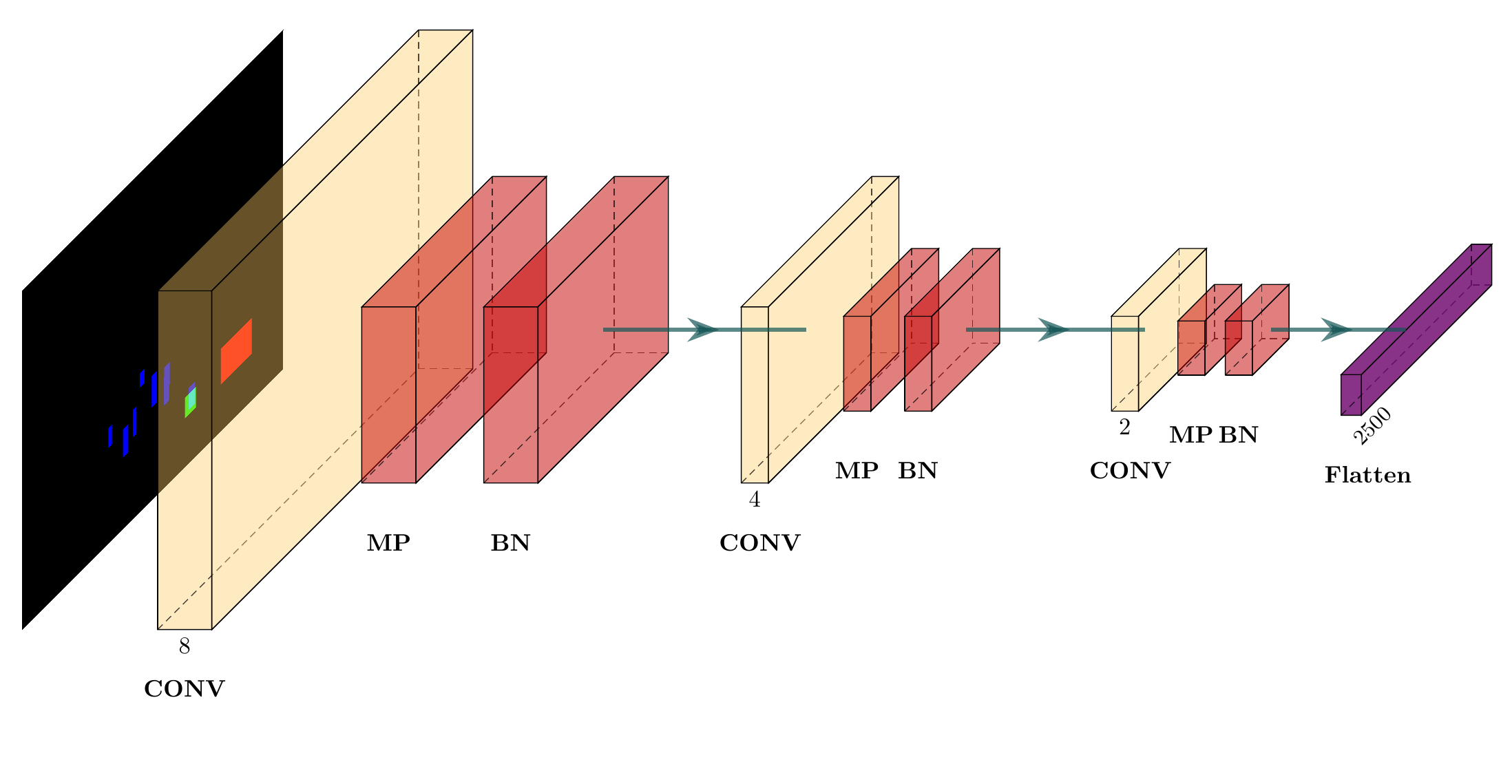}
	\caption[The CNN used for learning spatial features]{The CNN used for learning spatial features from an object frame. CONV stands for 2D convolutional layer, MP for Maximum Pooling layer and BN for Batch Normalization.}
	\label{fig:CVAEclassifierCNN}
\end{figure}

The X-Encoder manipulates two CNNs for learning spatial features from the object frame sequence and the optical--flow frame sequence, respectively. 
Without loss of generality, the object frame sequence using the sliding window (\eg~$\mathsf{w}=8$) method is taken as an example for explaining the learning process. 
First, each frame from the sliding window is passed to a CNN to learn spatial features. As shown in Fig.~\ref{fig:CVAEclassifierCNN}, the CNN has three 2D convolutional (CONV) layers with each one followed by a Maximum Pooling (MP) layer and a Batch Normalization (BN)~\cite{ioffe2015batch}. It takes the frame that contains object information as input and outputs a flattened feature vector. This process is done frame by frame for all the frames in the sliding window.
Then, the output feature vectors of all the frames are timely distributed as a sequence that maintains the same length as the window size, as shown in Fig.~\ref{fig:pipelineInteractionDetection} for the X-Encoder\footnote{This process works in the same way for the padding method with the predefined sequence length, instead of the sliding window size.}. The optical--flow frame sequence is processed by another CNN in a similar way to get the sequence of optical--flow feature vectors.
In the end, the object feature vectors and the optical--flow feature vectors are concatenated into a 2D feature vector as the final output of the X-Encoder.
Note that the CNN for the optical--flow frame sequence has a similar structure, except for the input channel number. The CNN for the object frame sequence has four channels that are dedicated to different road user types, whereas the CNN for the optical--flow frame sequence has three channels that are dedicated to the motion information (see Table.~\ref{tb:extractedfeatures}).

The Y-Encoder embeds the interaction label for each sequence. First, the sequence--wise label is replicated to align with the sequence length. Then, a fully connected (FC) layer is used to embed the replicated labels into a label vector. The original dimension of the label is only two after the one-hot encoding for the non-interaction and interaction classes, which is much smaller than the combined feature vector. The embedding increases the balance of the sizes of the label vector and the combined feature vector. The specific dimensionalities are shown in Fig.~\ref{fig:pipelineInteractionDetection}, which are hyper-parameters that can be changed in the experimental settings.

The prior Gaussian latent variables $\mathbf{z}$ are modulated by the encoded feature vector and the label vector from the X-Encoder and Y-Encoder, respectively. First, the outputs of the X-Encoder and Y-Encoder are concatenated along the time axis. Then, the concatenated features are passed to an FC layer and a following self-attention layer~\cite{vaswani2017attention}. The self-attention layer takes all the features along the time axis at the same time
and attentively learns the interconnections of the features globally. After that, an LSTM with two stacked hidden layers is used for learning the temporal features into a hidden state. In the end, the hidden state is fully connected by an FC layer and then split by two FC layers side by side. The two FC layers are trained to learn the mean and variance of the distribution of the latent variables $\mathbf{z}$, respectively.  

The Decoder is trained conditioned on the encoded feature vector from the X-Encoder and the latent variables. First, the encoded feature vector is concatenated with the latent variables and passed to an FC layer. Then, an LSTM with two stacked layers is used to learn the temporal dynamics. After that, two FC layers are used for fusion and dimension reduction. The Softmax activation function is added to the last FC layer for generating the probability of the interaction class at each frame. The output of the Decoder are the frame--wise predictions of the interaction class. In the end, the average voting scheme is used to summarize the frame--wise predictions to get the sequence--wise prediction for the interaction class. 

In inference time, the interactions for unseen vehicle turning sequences are classified using the trained CVAE model. First, the object and optical--flow information are encoded by the X-Encoder. A latent variable is sampled from the Gaussian distribution. Then the Decoder generates the probabilities of the interaction class for each sequence conditioned on the output of the X-Encoder and the sampled latent variable. The sampling is repeated multiple times at each step so that the Decoder generates diverse probabilities of the interaction class.  

\subsection{Estimation of uncertainty}
\label{subsec:interactionuncertainty}
Kernel density estimation (KDE)~\cite{parzen1962estimation,loftsgaarden1965nonparametric} is used to measure the uncertainty of the diverse predictions generated by the above multi-sampling process.
At each frame, the predictions $\{Y_{i, 1}^t,\,\cdots,Y_{i, N}^t\}$ are assumed to be i.i.d. samples drawn from an unknown density function $g(Y)$. $N$ is the total number of predictions and $t\leq T$, and $T$ is the total steps of the give sequence $i$. The KDE is calculated as:
\begin{equation}
    \hat{g}(Y)^t = \frac{1}{N}\sum_{i=1}^{N}K_h(Y-Y_i)=\frac{1}{Nh}\sum_{i=1}^{N}K\left(\frac{Y-Y_i}{h}\right),
\end{equation}
where $K(\cdot)$ is the Gaussian kernel function, $h$ is the smoothing parameter (also called \textit{bandwidth}). The log-likelihood of the average prediction at step $t$ is determined by $\mathcal{L}(\log(\hat{g}),\, \bar{Y_i})^t$, where $\bar{Y_i}$ is the average prediction. The uncertainty is defined as the residual of the normalized log-likelihood averaged over all the steps for the given sequence $i$, as denoted by Eq.~\eqref{eq:uncertainty}:
\begin{equation}
\label{eq:uncertainty}
    \Gamma_i = 1- \frac{1}{T}\sum_{t=0}^{T-1}\omega\mathcal{L}(\log(\hat{g}),\, \bar{Y_i})^t,
\end{equation}
where $\omega$ is the normalization parameter that scales the values to $[0, 1]$ and $\Gamma_i$ stands for the degree of uncertainty for the prediction over the sequence $i$.

\section{Data Acquisition and Pre-processing}
\label{sec:InteDetcData}

\begin{figure}[t!]
\centering
    \subfloat[The KoW left--hand intersection in Germany]{
    	\label{subfig:KoWscreenshot}
    	\begin{minipage}{3.5in}
    		\centering
    		\includegraphics[trim=0in 0in 0in 2.5in, clip=true, width=3.45in]{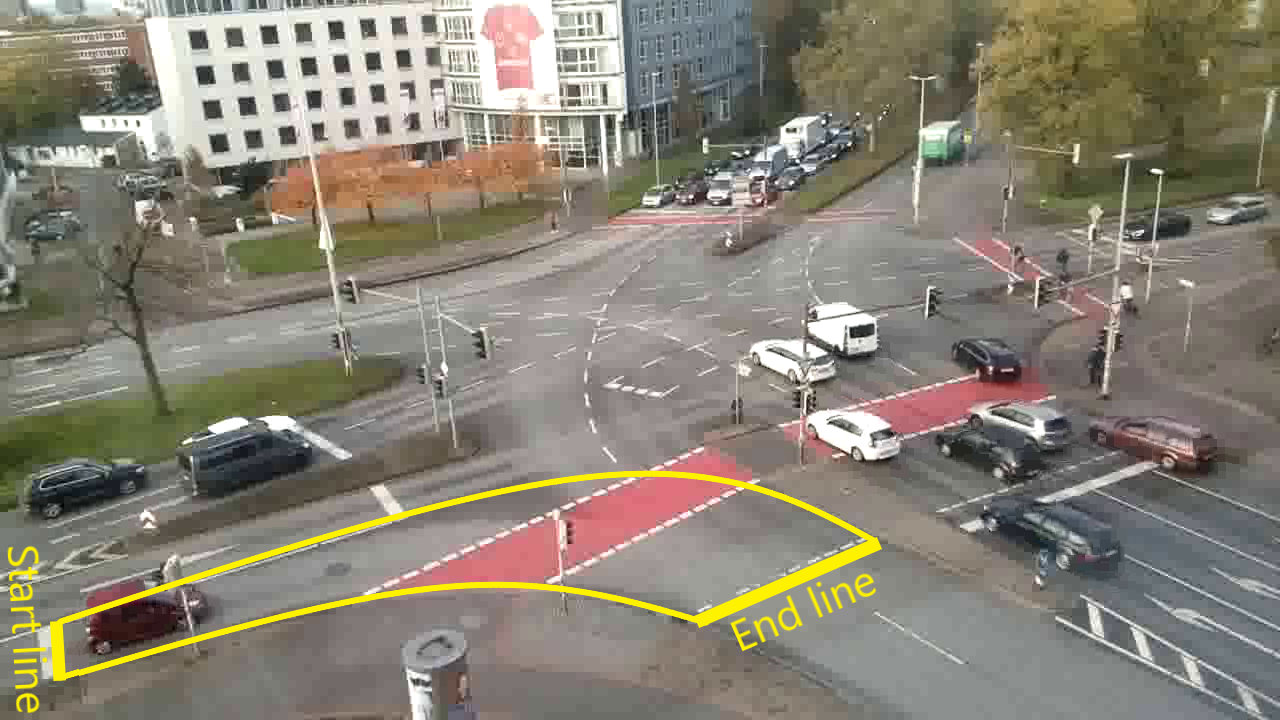}
    	\end{minipage}
    }%
    
    \subfloat[The NGY right--hand intersection in Japan]{
    	\label{subfig:NGYscreenshot}
    	\begin{minipage}{3.5in}
    		\centering
    		\includegraphics[trim=0in 2in 0in 2.8in, clip=true, width=3.45in]{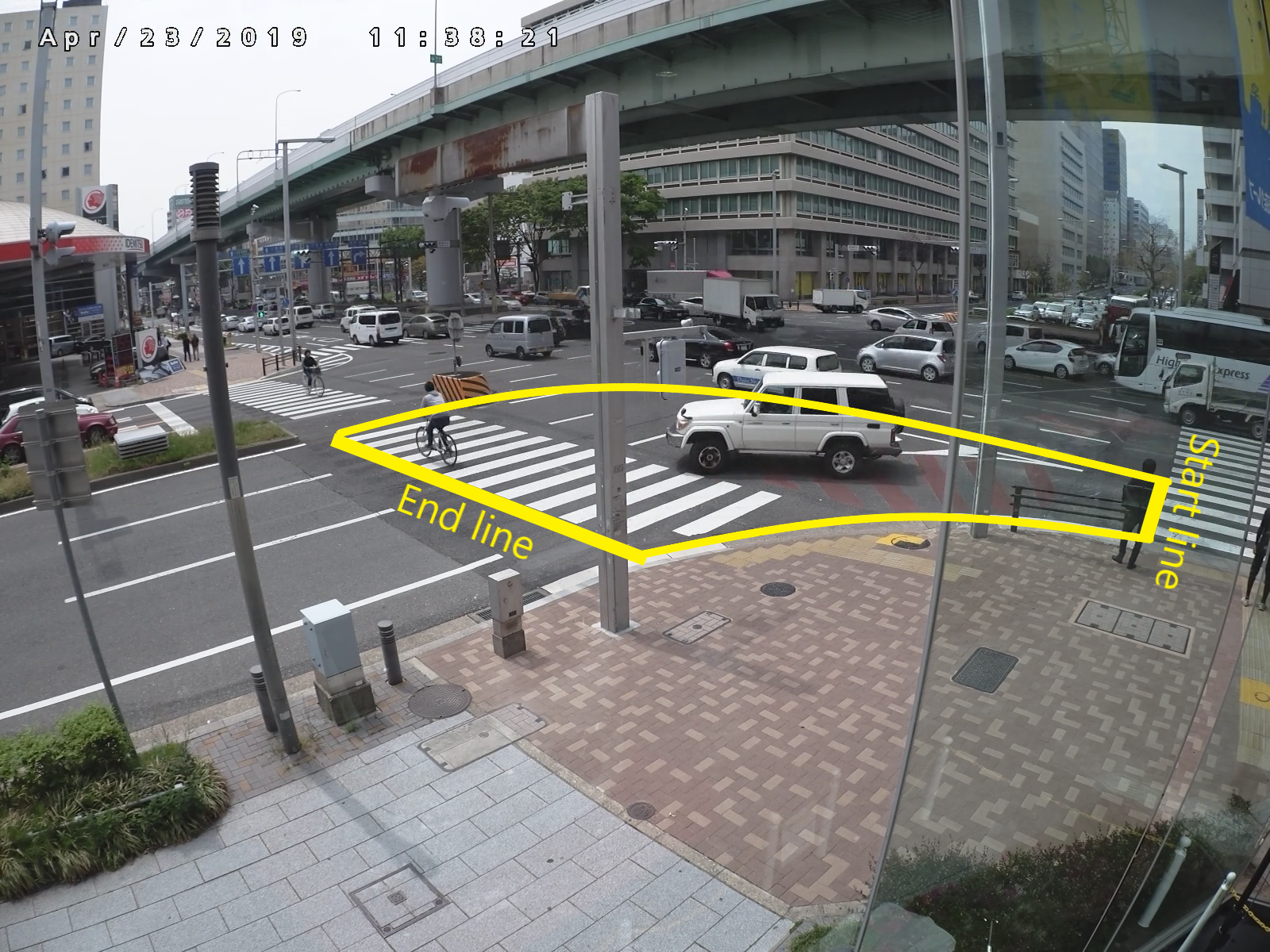}
    	\end{minipage}
    }%
\caption[Screenshots of two intersections]{The screenshots of KoW and NGY intersections. Vehicle turning sequences are constrained in the yellow contours.}
\label{fig:intersections}
\end{figure}

Real--world datasets were acquired to test the performance of the proposed model for interaction detection. Fig.~\ref{fig:intersections} shows the screenshots of the two intersections where various traffic scenes were recorded. The KoW dataset was acquired by \cite{koetsier2019trajectory} from a very busy right--turn intersection in a German city. The videos recorded traffic conditions from 00:02~a.\,m. to 11:58~p.\,m. on November 8th and 9th, 2019 in Hannover. The videos were recorded in $1280 \times 720$ pixels at \SI{25}{fps} by a camera module (Raspberry Pi Camera Module v2) installed inside a building (ca. \SI{20}{m} ground elevation) facing the intersection and stored in .h264 format. We use an approximately 14-hour sub-footage from two seven--hour segments (8~a.\,m. to 3~p.\,m. on both the 8th and 9th), when there was enough traffic and adequate ambient light to perform stable image processing for feature extraction. The NGY dataset was provided by Nagoya Toyopet Corporation. It was acquired from an extremely busy left--turn intersection in a Japanese city. In total, approximately 24 hours of traffic footage from an oblique view at one of the major intersections in Nagoya were recorded from 11~a.\,m. to 11~a.\,m. on April 23rd and 24th, 2019. The videos were recorded in $1600\times1200$ pixels at \SI{30}{fps} using a camera (Panasonic WV-SF781L) installed inside a building (ca. \SI{3}{m} ground elevation) adjacent to the intersection and stored in .mp4 format. Similarly, we use a twelve-hour sub-footage recorded from 11~a.\,m. to 6~p.\,m. on the 23th and from 6~a.m. to 11~a.m. on the 24th.

Both datasets were pre-processed for later usage. Due to the missing camera intrinsic and extrinsic parameters, no projection was done for extracting the trajectory data. The pre-process aimed to identify vehicle turning sequences and extract all the road users' type, position and motion information. First, two annotators for each dataset manually detected the sequence scenes where a vehicle turned right at the KoW intersection or left at the NGY intersection, and extracted the time intervals of the vehicle staying in the yellow contours (see Fig.~\ref{fig:intersections}). 
The annotators independently determined whether or not interactions occurred in each scene, afterwards they revised and agreed with each other\footnote{Less than \SI{1}{\percent} of the sequences were initially annotated differently.} and labeled each scene as ``non-interaction'' or ``interaction''. Then, YOLOv3~\cite{redmon2016you} and M2Det~\cite{zhao2019m2det} were used to detect all the traffic related objects at the original frame rate of the KoW (\SI{25}{fps}) and NGY (\SI{30}{fps}) datasets, respectively. Note that these two sources of data were from different providers so that the camera settings and the object detection algorithms were not unified. Considering the change between two consecutive frames is small, a possibly failed detection in the current frame is supplemented by the detection in the previous or the next frame if either of them was available. Otherwise, the sequences with failed detection and no supplementation were discarded. 
In addition, the dense optical--flow algorithm~\cite{farneback2003two} was used to extract the optical--flow information from the sequences. Different from the object detection, the frame rate was down--sampled to the half of the original frame rate, \ie~\SI{12.5}{fps} for KoW and \SI{15}{fps} for NGY.
This aims to reduce the computational cost and increase the offset of moving pixels between two consecutive frames, so as to improve the extraction performance~\cite{farneback2003two} of optical flow. In the end, both the object and the optical--flow sequences were aligned with the down--sampled frame rate in each dataset, which is used as the time step for interaction detection. 

\begin{figure}[t!]
\centering
    \subfloat[]{
    	\label{subfig:seq-nonvsin}
    	\begin{minipage}{0.236\textwidth}
    		\centering
    		\includegraphics[trim=0in -0.1in 0in 0in, width=\textwidth]{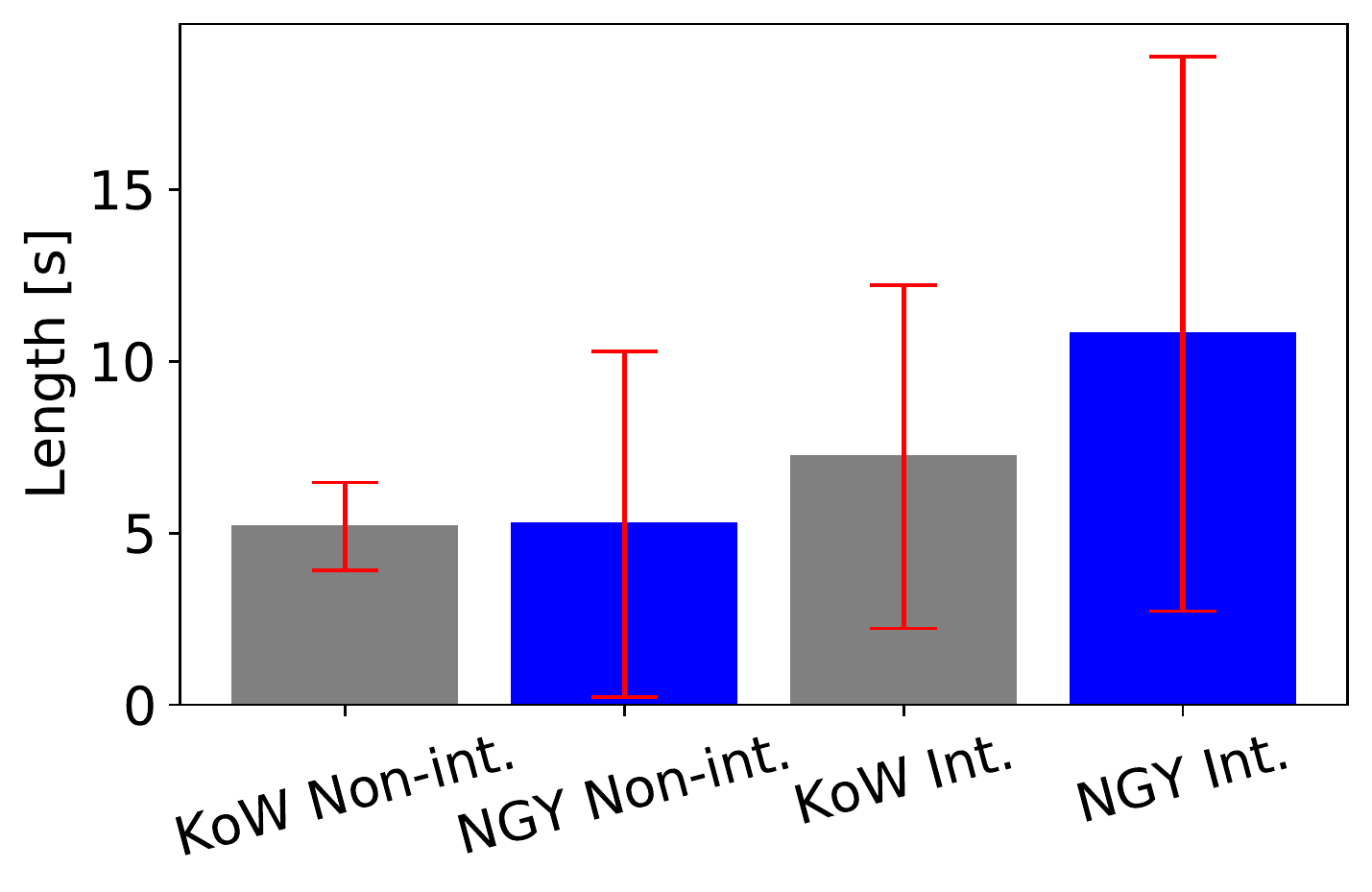}
    	\end{minipage}%
    }%
    \subfloat[]{
    	\label{fig:seq-frame-overdata}
    	\begin{minipage}{0.236\textwidth}
    		\centering
    		\includegraphics[trim=0in 0in 0in 0in, clip=true, width=\textwidth]{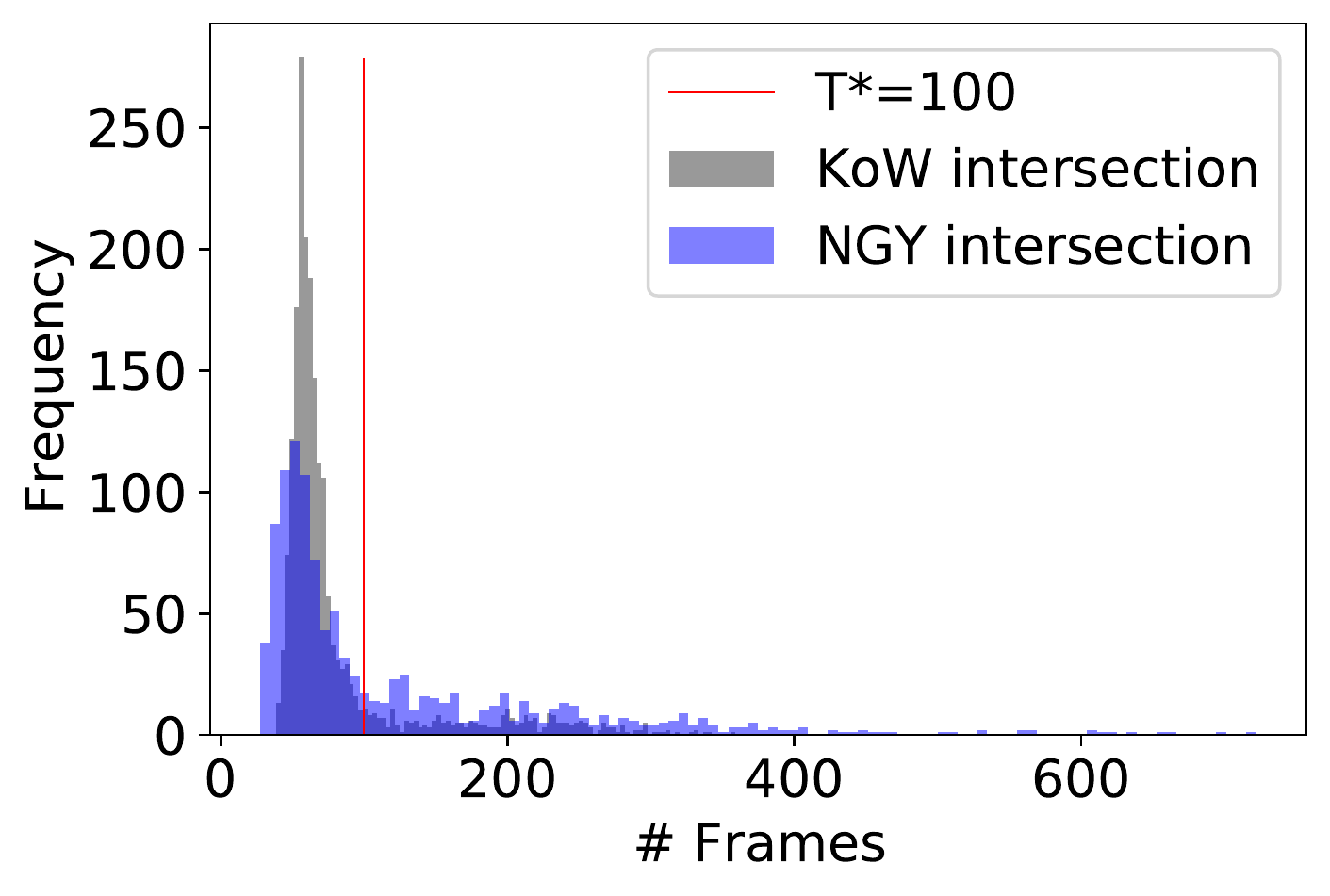}
    	\end{minipage}%
    }
\caption[Sequence lengths in the KoW and NGY datasets]{Sequence lengths in the KoW and NGY datasets. (a) Standard deviation is denoted by the red error bar. (b) Sequence length measured by the number of frames and $T^*$ is the length threshold for the padding method.}  
\label{fig:seqdistrubitions}
\end{figure}
The data processing yields over 2000 vehicle turning sequences with varying lengths, as denoted by Fig.~\ref{fig:seqdistrubitions}.
Within each dataset, sequence lengths measured in seconds are very different.
The non-interaction sequences are significantly shorter than the interaction ones (U-test, $U=382142$, $p\ll0.01$ for KoW and $U=67199$,~$p\ll0.01$ for NGY), and the standard deviation in each class in both datasets deviates in a large range. 
This indicates that the duration of a sequence is not an accurate feature for the detection task; a short sequence duration does not necessarily imply no interaction. In addition, the sequence lengths of all the sequences over each dataset are very unevenly distributed, \ie~long--tail distributed, especially for the NGY dataset, see Fig.~\ref{fig:seq-frame-overdata}.
Across the datasets, the sequences in the KoW and NGY datasets are different in terms of not only travel direction but also frame size and rate, as well as sequence length in general. 
Though non-interaction sequences from both datasets have similar sequence lengths, \ie~on average non-interaction sequences have a length of \SI{5.2}{s} in KoW and \SI{5.3}{s} in NGY, interaction sequences in NGY have a longer average sequence length (\SI{10.8}{s}) than the ones in KoW (\SI{7.2}{s}). 
Due to the higher density of traffic at the NGY intersection compared to the KoW intersection, vehicles often had to wait for more pedestrians and cyclists crossing. The above differences make cross dataset validation very difficult (more details in Sec.~\ref{subsec:crossdatavalidation}). 

\begin{table}[hbpt!]
\caption[Video frame sequences used for interaction detection]{Video frame sequences used for interaction detection. Sequence length is measured by frame numbers and the sample sizes of the classes were balanced for each sets.}
\setlength{\tabcolsep}{3pt}
\centering
\begin{tabular}{ccccccc}
\toprule
Name & Input form   & Max. length$^{*}$ & Training & Validation & Test    & Total   \\ \hhline{=======}
KoW         & sliding & 500              & 360/360  & 90/90      & 192/192 & 642/642 \\
KoW         & padding      & 100              & 352/352  & 88/88      & 188/188 & 628/628 \\
NGY         & sliding & 500              & 291/291  & 74/74      & 159/159 & 530/530 \\
NGY         & padding      & 100              & 132/132  & 33/33      & 70/70   & 235/235 \\ \bottomrule
\end{tabular}
\label{tb:interactiondatapartition}
\end{table}

The acquired datasets were further prepared for training the detection models, which involves sample balancing, sequence padding and dataset partitioning.
The number of samples in each class was balanced to perform unbiased training. For both datasets, the maximum number of sequences in each class was set to a value that is close to the capacity of the smaller class. Note that the small amount of very long sequences (\ie~$>500$ frames,  see Fig.~\ref{fig:seq-frame-overdata}) were not used for the experiments. All such sequences are from the class of interaction, \ie~vehicles had to wait for a long time to let other road users cross the intersection. The removal of these sequences balances sample size and length in both classes, in order to prevent a model being biased towards the interaction class.
Sequences with a smaller number of frames than the threshold $T^\ast$ are padded with zeros for the padding method (see Sec.~\ref{subsec:interactionproblemformulation}).
However, if $T^\ast$ is too large most of the sequences will be padded with zeros and this will lead to noisy samples; if $T^\ast$ is too small many long sequences will be excluded.
To balance the trade-off, based on the sequence length distributions over the datasets (Fig.~\ref{fig:seq-frame-overdata}),  $T^\ast$ was set to 100 frames for both datasets so that the majority of all the sequences were included. The sequences shorter than or equal to $T^\ast$ 
were preserved for the experiments of the models that use the padding method. Sequences longer than $T^\ast$ exceed the maximum length of the input size that the models can handle. Therefore, they were discarded.
Under the balanced criteria above for each class, both the datasets were then randomly split into training and test sets by the ratio of $70:30$. Additionally $20\,\%$ of the training data was separated as an independent validation set to monitor the process of training. Table~\ref{tb:interactiondatapartition} lists the statistics of the final data used for the experiments after the preparation steps.

\section{Experiments}
\label{sec:InteDetcExperimentSettings}

\subsection{Baseline and ablative models}
\label{subsec:interactionbaseline}
To evaluate the performance of the proposed CVAE model, it is compared with a baseline model. The baseline model is a sequence-to-sequence encoder--decoder model that uses the same input features from the object and motion information for interaction detection~\cite{cheng2020automatic}. It has the same structure as the X-Encoder and the Decoder that are implemented in the CVAE model (Fig.~\ref{fig:pipelineInteractionDetection}). The difference between these two models is the sample generation process. The baseline model is a discriminative model and does not use the class label information and the conditional information for learning the latent variables that mimic the stochastic behavior in vehicle--VRU interactions. Without randomly sampling from the Gaussian latent variables, the output of the sequence-to-sequence encoder--decoder model is deterministic. 

A series of ablative models are designed to analyze the contribution of the object information (\textit{ob}), the optical--flow information (\textit{op}), and the self-attention mechanism (\textit{att}). The ablative models are trained by removing one of the aforementioned parts, as denoted in Table~\ref{tb:interactionclassifiers}. 

\begin{table}[hbpt!]
\caption[The models with different input structures]{The models with different input structures.}
\label{tb:interactionclassifiers}
\centering
\begin{tabular}{lcccc}
\toprule
Model name                     & (\textit{ob})      & (\textit{op})           & (\textit{att}) & Sample generation \\ \hhline{=====}
\textit{[S+ob+op+att]}$^1$   & $\surd$            & $\surd$                  & $\surd$        & -                 \\
\textit{[C+op+att]}         & -                  & $\surd$                  & $\surd$        & $\surd$           \\
\textit{[C+ob+att]}         & $\surd$            & -                        & $\surd$        & $\surd$           \\
\textit{[C+ob+op]}          & $\surd$            & $\surd$                  & -              & $\surd$           \\
\textit{[C+ob+op+att]}$^2$  & $\surd$            & $\surd$                  & $\surd$        & $\surd$           \\
\bottomrule   
\end{tabular}
\begin{tabular}{@{}c@{}}
\multicolumn{1}{p{\textwidth}}{$^{1}$the baseline model; $^{2}$the complete CVAE model} \\
\end{tabular}
\end{table}

\subsection{Evaluation metrics}
\label{subsec:interactionevaluationmetrics}
Tested samples are categorized according to the comparison between their ground truth and the predicted labels, as listed in Table~\ref{tb:evaluationcategory}. Accuracy, Precision, Recall and F1-score are applied to measure the performance of interaction detection on the test data from both the KoW and NGY intersections. 

\begin{table}[hbpt!]
\caption[Categories of tested samples]{Categories of tested samples}
\centering
\begin{tabular}{lll}
\toprule
Category name               & Ground truth & Prediction \\ \hhline{===}
TP: true positive  & interaction & interaction   \\
TN: true negative  & non-interaction & non-interaction  \\
FP: false positive & non-interaction & interaction  \\
FN: false negative & interaction & non-interaction  \\ \bottomrule
\end{tabular}
\label{tb:evaluationcategory}
\end{table}

Accuracy is the fraction of the number of the correctly predicted samples over the total number of samples.
\begin{equation*}
    \text{Accuracy}=(\text{TP}+\text{TN})/({\text{TP}+\text{TN}+\text{FP}+\text{FN}})\,.
\end{equation*}

Precision is the fraction of the number of the TP samples over the number of predicted positive samples.
\begin{equation*}
    \text{Precision}={\text{TP}}/({\text{TP}+\text{FP}})\,.
\end{equation*}

Recall is the fraction of the number of the TP samples over the number of actual positive samples in the whole dataset. 
\begin{equation*}
    \text{Recall}={\text{TP}}/({\text{TP}+\text{FN}})\,.
\end{equation*}

F1-score is used to provide a measurement of the overall performance of a model. It is defined as the so-called harmonic mean of precision and recall.
\begin{equation*}
\label{eq:f1score}
    \text{F1-score} = 2\times({\text{Precision}\cdot\text{Recall}})/({\text{Precision}+\text{Recall}})\,.
\end{equation*}

\subsection{Experimental Settings}
\label{subsec:settings}
The kernel size of the CNNs in each layer is set to 8, 4, and 2 respectively with a stride of 2 and the same padding for the borders. 
The size of the first hidden layer of the LSTM is set to 64 and the second hidden layer is 32. 
The size of the latent variables is set to 64. 
All the models are trained by a learning rate of $10^{-4}$ with a zero decay using the Adam optimizer ($\beta_1=0.9$ and $\beta_2=0.999$)~\cite{DBLP:journals/corr/KingmaB14}.
The batch size is set to 32, and all the models were trained for 50 epochs on an NVIDIA Quadro T2000 GPU.
In inference time, the number of sampling $N$ is set to 100 for all the CVAE--based models.   

\section{Results}
\label{sec:InteDetcResults}
This section presents the quantitative and qualitative results for each intersection, as well as the discussion of the results. 

\subsection{Quantitative results}
\label{sec:InteDetcResults-quantitative}
The quantitative results are summarized in Table~\ref{tb:KoW-results} and \ref{tb:NGY-results} for the right--turn KoW intersection and the left--turn NGY intersection, respectively. Due to the multi--sampling process the results of the CVAE--based models are not deterministic, hence the corresponding standard deviations are provided. 

Table~\ref{tb:KoW-results} shows the results of the interaction detection at the right--turn intersection. (1) Both the sliding window and padding methods yield similar and very accurate results for interaction detection using the combined information from object detection and optical flow. The accuracy and F1-score are both above 0.95. (2) Compared to the baseline models \textit{[S2S+ob+op+att]}, the proposed models \textit{[CVAE+ob+op+att]} have a slightly better performance using the sliding window method and comparable performance using the padding method. (3) Compared to the ablative models, the combined information improves the performance using the sliding window method. However, the improvement is not obvious using the padding method, especially compared to the ablative model that only uses the object information. On the other hand, regardless of the sliding window method or padding method, the ablative models that merely use optical--flow information only achieve an accuracy below 0.70. (4) The self-attention mechanism does not lead to an obviously better or worse performance using either the sliding window or padding method.

\begin{table}[hbpt!]
\caption[Detection results of a right--turn intersection]{Detection results of the right--turn intersection on the KoW dataset. Best values are highlighted in boldface.}
\centering
\setlength{\tabcolsep}{1.4pt}
\begin{tabular}{llllll}
\toprule
Model                   & shape   & Accuracy                  & Precision                 & Recall                    & F1-score  \\ \hhline{======}
\textit{[S+ob+op+att]}  & sli. & 0.951                     & 0.935                     & 0.969                     & 0.951                          \\ 
\textit{[C+op+att]}     & sli. & 0.692$_{\pm0.006}$          & 0.717$_{\pm0.007}$          & 0.635$_{\pm0.011}$          & 0.673$_{\pm0.008}$               \\
\textit{[C+ob+att]}     & sli. & 0.952$_{\pm0.002}$          & 0.934$_{\pm0.002}$          & \textbf{0.973$_{\pm0.004}$} & 0.953$_{\pm0.002}$               \\
\textit{[C+ob+op]}      & sli. & \textbf{0.965$_{\pm0.001}$} & \textbf{0.976$_{\pm0.003}$} & 0.953$_{\pm0.0}$             & \textbf{0.964$_{\pm0.001}$}      \\
\textit{[C+ob+op+att]}  & sli. & 0.961$_{\pm0.002}$          & 0.969$_{\pm0.004}$          & 0.953$_{\pm0.0}$             & 0.961$_{\pm0.002}$               \\ \midrule
\textit{[S+ob+op+att]}  & pad.      & 0.963                     & 0.944                     & 0.984                     & 0.964                          \\ 
\textit{[C+op+att]}     & pad.      & 0.610$_{\pm0.008}$          & 0.649$_{\pm0.011}$          & 0.479$_{\pm0.012}$          & 0.551$_{\pm0.010}$               \\
\textit{[C+ob+att]}     & pad.      & \textbf{0.967$_{\pm0.002}$} & \textbf{0.955$_{\pm0.003}$} & 0.980$_{\pm0.003}$          & \textbf{0.967$_{\pm0.002}$}      \\
\textit{[C+ob+op]}      & pad.      & 0.966$_{\pm0.003}$          & 0.946$_{\pm0.004}$          & \textbf{0.989$_{\pm0.001}$} & \textbf{0.967$_{\pm0.002}$}      \\
\textit{[C+ob+op+att]}  & pad.      & 0.962$_{\pm0.002}$          & 0.952$_{\pm0.003}$          & 0.973$_{\pm0.0}$             & 0.963$_{\pm0.002}$               \\ \bottomrule
\end{tabular}
\label{tb:KoW-results}
\end{table}

Table~\ref{tb:NGY-results} shows the results of the interaction detection at the left--turn intersection. (1) Both the sliding window and padding methods yield reasonable results for interaction detection using the combined information. However, the predictions of the sliding window method are more accurate than the padding method. (2) Compared to the baseline models, the CVAE models using the combined information achieve better performance, especially by using the sliding window method (\eg~ about 0.05 increment in F1-score). (3) Compared to the ablative models, the improvement by using the combined information can be found in both the sliding window and padding methods. (4) The best performance, especially measured by recall (0.916) and F1-score (0.892) on the NGY dataset, is achieved by the proposed CVAE model using the sliding window method with the self-attention mechanism. 

\begin{table}[hbpt!]
\caption[Detection results of a left--turn intersection]{Detection results of the left--turn intersection on the NGY dataset. Best values are highlighted in boldface.}
\centering
\setlength{\tabcolsep}{1.4pt}
\begin{tabular}{llllll}
\toprule
Model                     & Shape   & Accuracy                  & Precision                 & Recall                    & F1-score                         \\ \hhline{======}
\textit{[S+ob+op+att]} & sli. & 0.849                        & 0.878                        & 0.811                     & 0.843                            \\ 
\textit{[C+op+att]}    & sli. & 0.878$_{\pm0.004}$          & 0.854$_{\pm0.004}$          & 0.912$_{\pm0.006}$          & 0.882$_{\pm0.004}$               \\
\textit{[C+ob+att]}    & sli. & 0.734$_{\pm0.008}$          & 0.698$_{\pm0.007}$          & 0.824$_{\pm0.009}$          & 0.756$_{\pm0.007}$               \\
\textit{[C+ob+op]}     & sli. & 0.882$_{\pm0.006}$          & \textbf{0.915$_{\pm0.004}$} & 0.842$_{\pm0.009}$          & 0.887$_{\pm0.006}$               \\
\textit{[C+ob+op+att]} & sli. & \textbf{0.889$_{\pm0.005}$} & 0.869$_{\pm0.005}$          & \textbf{0.916$_{\pm0.007}$} & \textbf{0.892$_{\pm0.004}$}      \\ \midrule
\textit{[S+ob+op+att]} & pad. & 0.721                        & 0.712                        & 0.743                     & 0.727                            \\ 
\textit{[C+op+att]}    & pad. & 0.764$_{\pm0.010}$          & 0.756$_{\pm0.011}$          & 0.808$_{\pm0.013}$          & 0.781$_{\pm0.009}$               \\
\textit{[C+ob+att]}    & pad. & 0.683$_{\pm0.012}$          & 0.661$_{\pm0.011}$          & 0.751$_{\pm0.019}$          & 0.703$_{\pm0.013}$               \\
\textit{[C+ob+op]}     & pad. & \textbf{0.782$_{\pm0.007}$} & \textbf{0.763$_{\pm0.010}$} & \textbf{0.819$_{\pm0.014}$} & \textbf{0.790$_{\pm0.009}$}      \\
\textit{[C+ob+op+att]} & pad. & 0.742$_{\pm0.007}$          & 0.750$_{\pm0.009}$          & 0.728$_{\pm0.010}$          & 0.739$_{\pm0.007}$               \\ \bottomrule
\end{tabular}
\label{tb:NGY-results}
\end{table}

\begin{figure}[t!]
\centering
    \subfloat[KoW sliding window]{
    	\label{subfig:uncertainty-Kow-sliding}
    	\hspace{-0.2cm}\begin{minipage}{0.25\textwidth}
    		\centering
    		\includegraphics[trim=0in 0in 0in 0in, width=\textwidth]{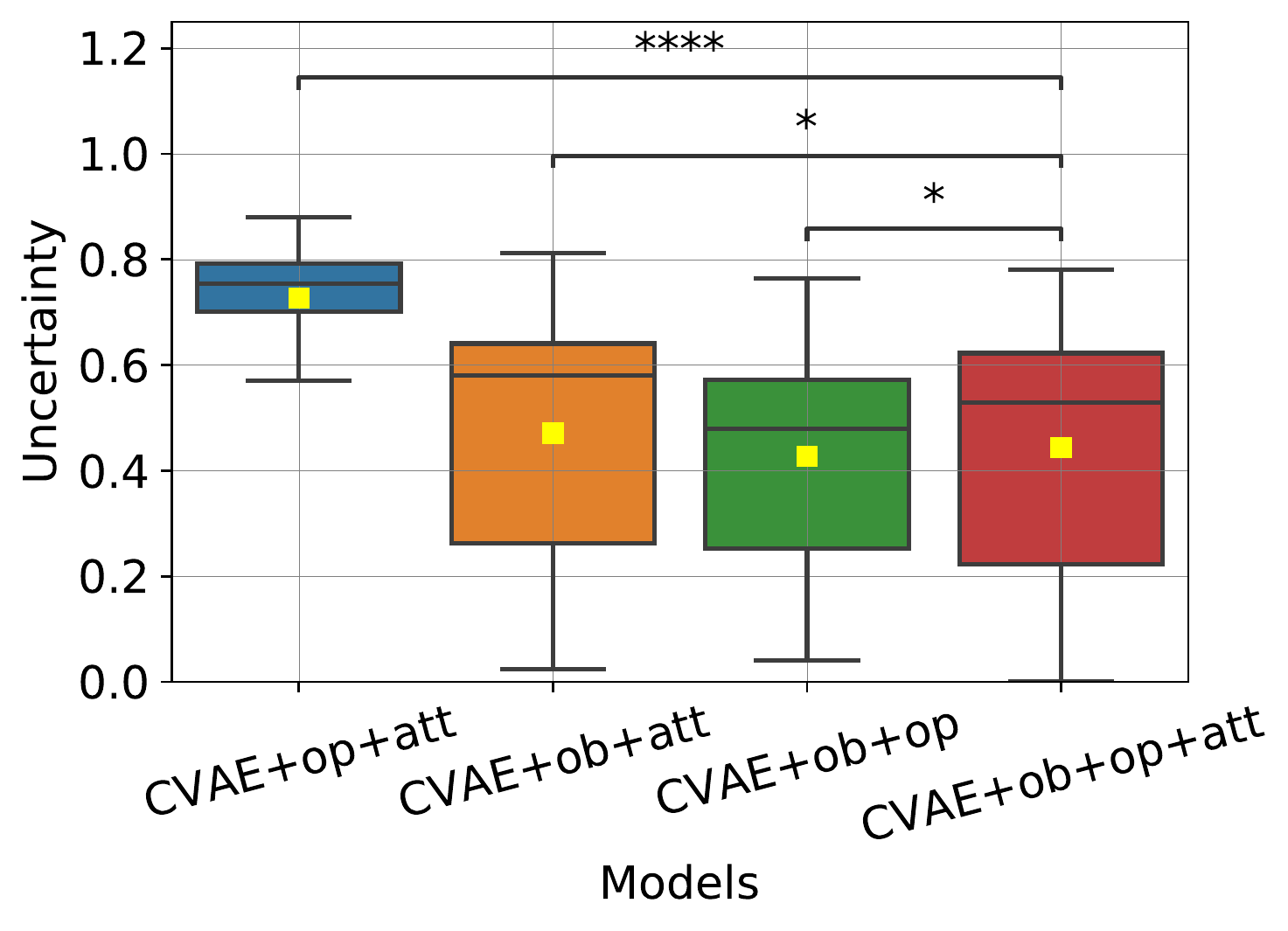}
    	\end{minipage}%
    }%
    \subfloat[KoW padding]{
    	\label{subfig:uncertainty-Kow-pad}
    	\hspace{-0.2cm}\begin{minipage}{0.25\textwidth}
    		\centering
    		\includegraphics[trim=0in 0in 0in 0in, width=\textwidth]{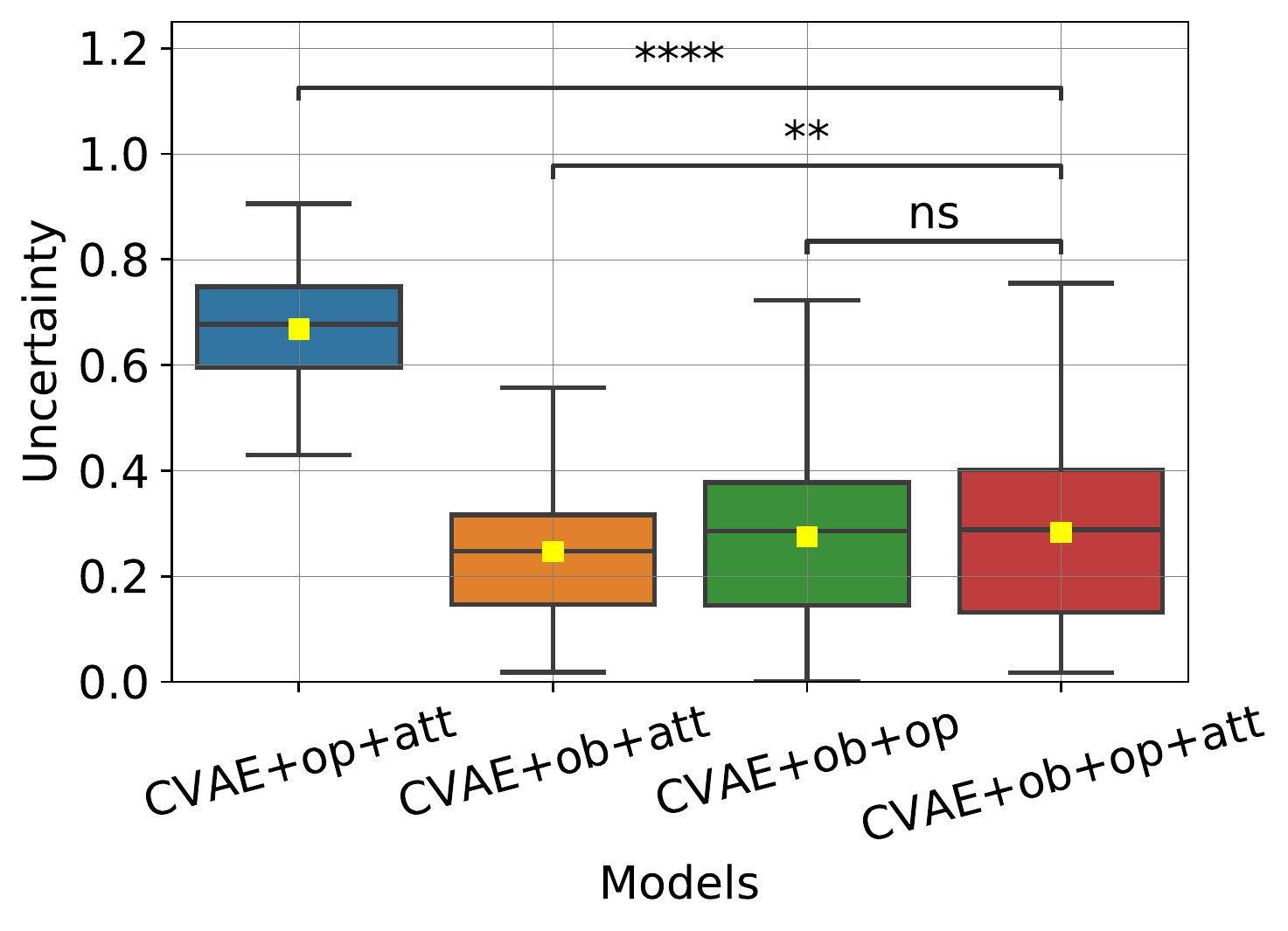}
    	\end{minipage}%
    }%
    \\
    \subfloat[NGY sliding window]{
    	\label{subfig:uncertainty-NGY-sliding}
    	\hspace{-0.2cm}\begin{minipage}{0.25\textwidth}
    		\centering
    		\includegraphics[trim=0in 0in 0in 0in, width=\textwidth]{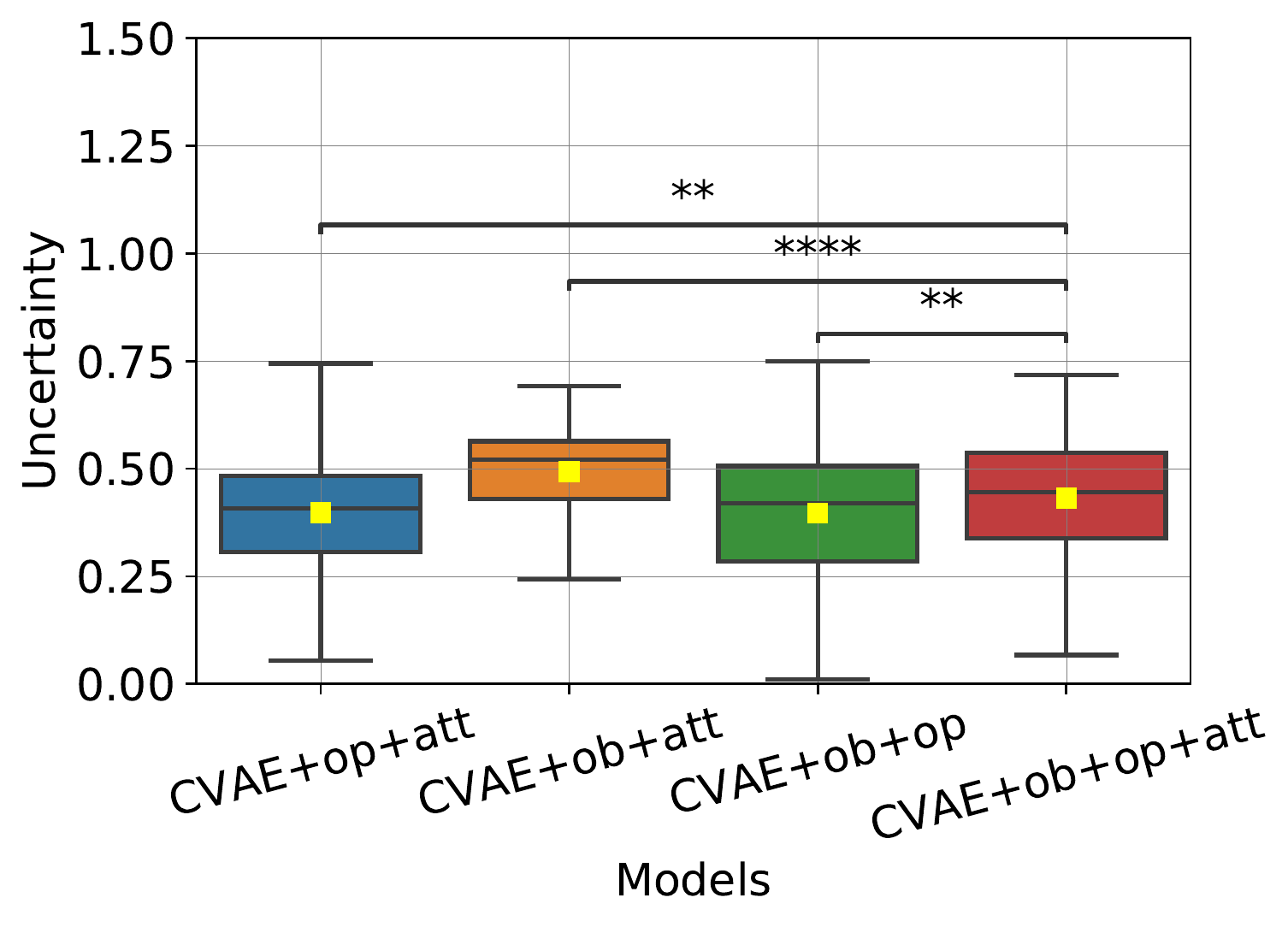}
    	\end{minipage}
    }%
    \subfloat[NGY padding]{
    	\label{subfig:uncertainty-NGY-pad}
    	\hspace{-0.2cm}\begin{minipage}{0.25\textwidth}
    		\centering
    		\includegraphics[trim=0in 0in 0in 0in, width=\textwidth]{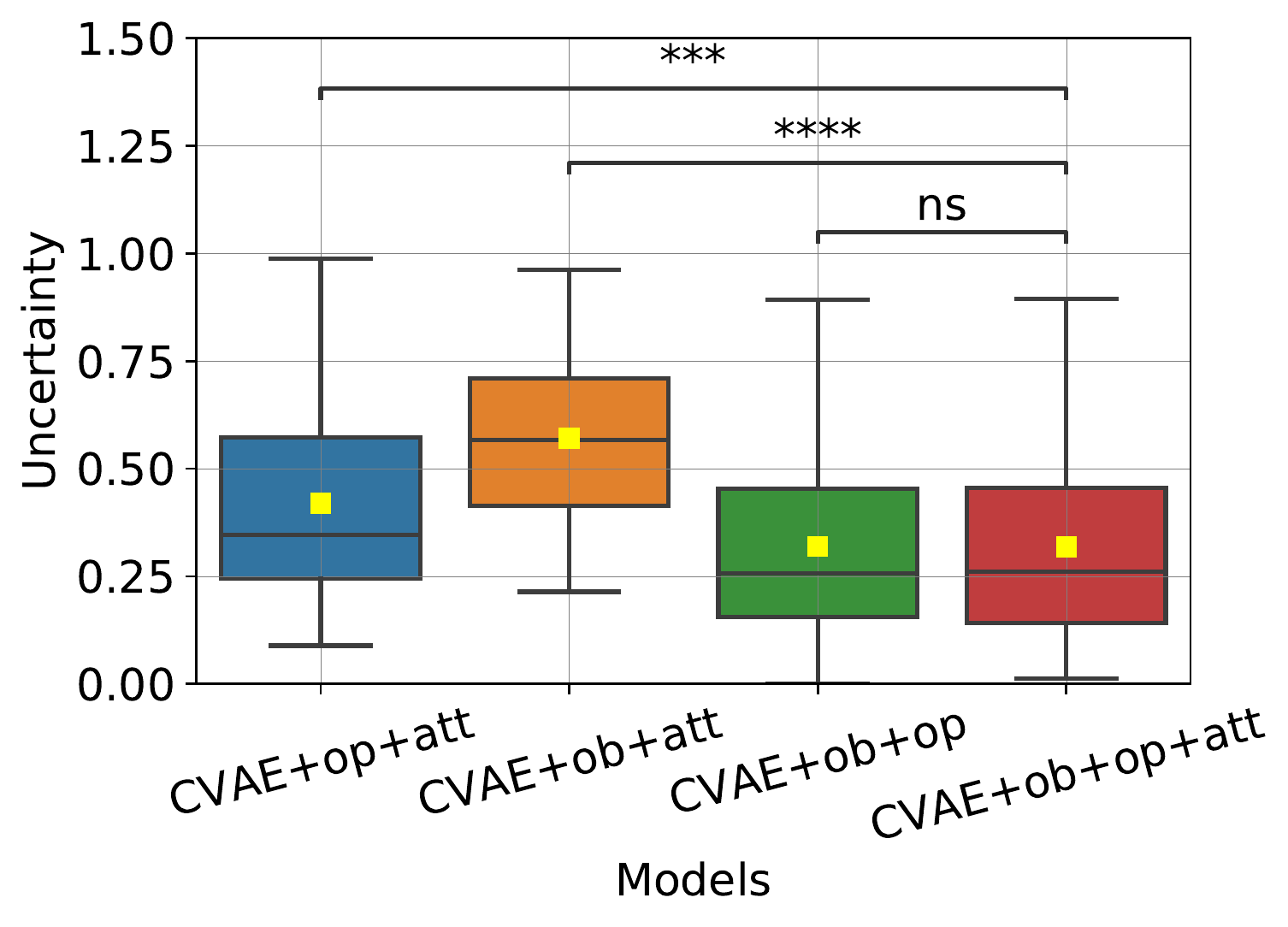}
    	\end{minipage}%
    }%
\caption[Uncertainty measurement of the CVAE--based models]{Uncertainty measure of the CVAE--based models tested on the KoW and NGY datasets. The mean value is denoted by the yellow square in each box-plot. The uncertainty levels across the models are compared using the Mann-Whitney U-test. p-values are annotated using * or \text{ns} (not significant), where ns: \small{$0.05 < p <= 1.00$, *: $10^{-2} < p <= 0.05$, **: $10^{-3} < p <= 10^{-2}$, ***: $10^{-4} < p <= 10^{-3}$, and ****: $p <= 10^{-4}$}.}  
\label{fig:uncertainty-cvae}
\end{figure}

The Kernel Density Estimation (KDE) function (see Sec.~\ref{subsec:interactionuncertainty}) is used to measure the uncertainty levels of the CVAE--based models with different input structures. The uncertainties of the CVAE--based models are plotted in Fig.~\ref{fig:uncertainty-cvae} and compared by the Mann-Whitney U-test. Fig.~\ref{subfig:uncertainty-Kow-sliding} and \ref{subfig:uncertainty-Kow-pad} demonstrate that the CVAE--based models \textit{[CVAE+op+att]} using only the optical--flow information generate significantly more uncertain predictions than the other models tested on the KoW dataset. 
This pattern is consistent with the prediction performance that they also yield less accurate predictions. 
A similar pattern can be observed from the CVAE--based models \textit{[CVAE+ob+att]} using only the object information (Fig.~\ref{subfig:uncertainty-NGY-sliding} and \ref{subfig:uncertainty-NGY-pad}) tested on the NGY dataset. When the uncertainty level is high in the predictions, the accuracy level also drops. 

\begin{figure}[t!]
\centering
    \subfloat[KoW sliding window]{
    	\label{subfig:cm-Kow-sliding}
    	\hspace{-0.2cm}\begin{minipage}{0.24\textwidth}
    		\centering
    		\includegraphics[trim=0in 0in 0in 0in, width=\textwidth]{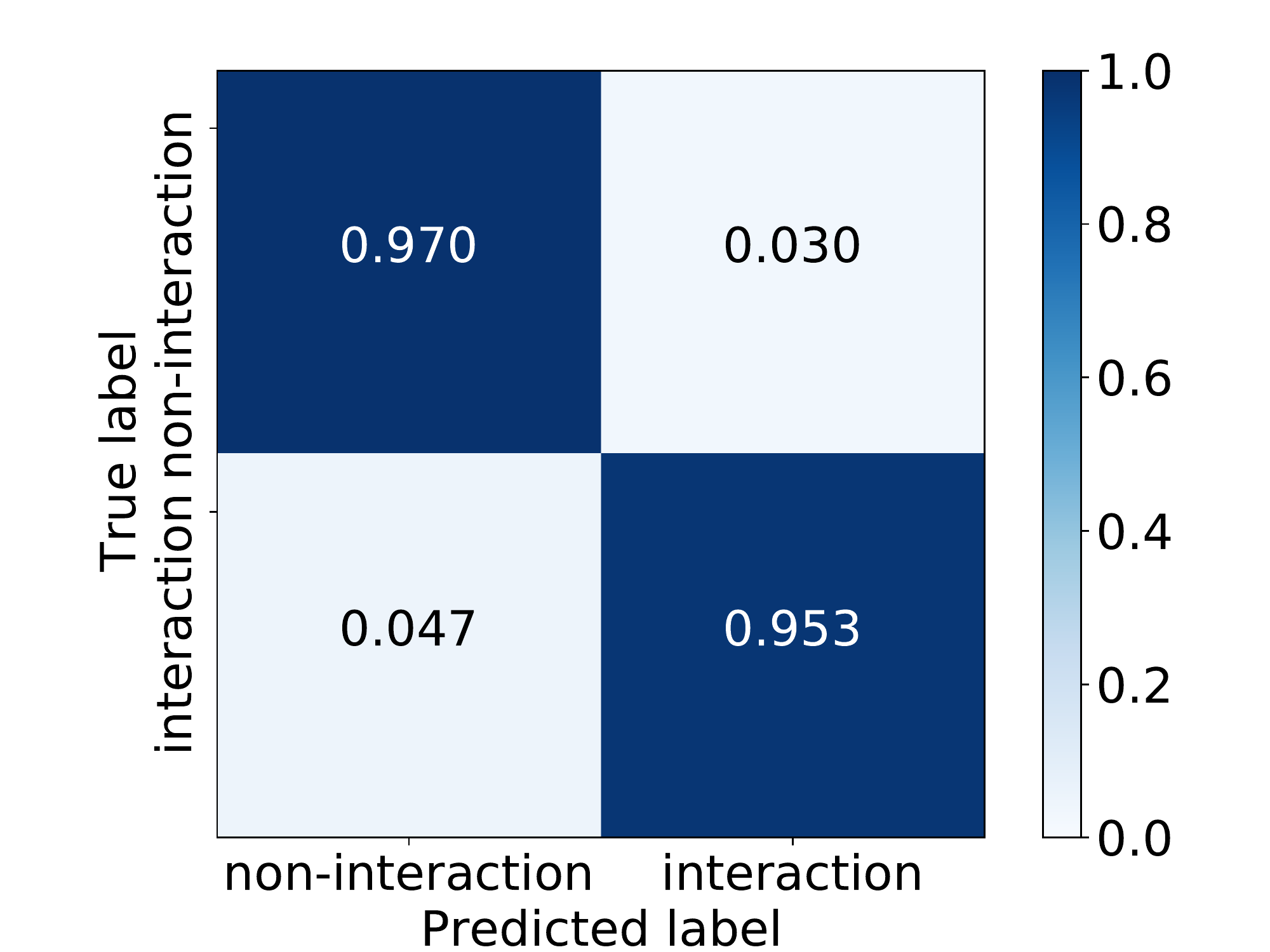}
    	\end{minipage}
    }
    \subfloat[KoW padding]{
    	\label{subfig:cm-Kow-pad}
    	\hspace{-0.2cm}\begin{minipage}{0.24\textwidth}
    		\centering
    		\includegraphics[trim=0in 0in 0in 0in, width=\textwidth]{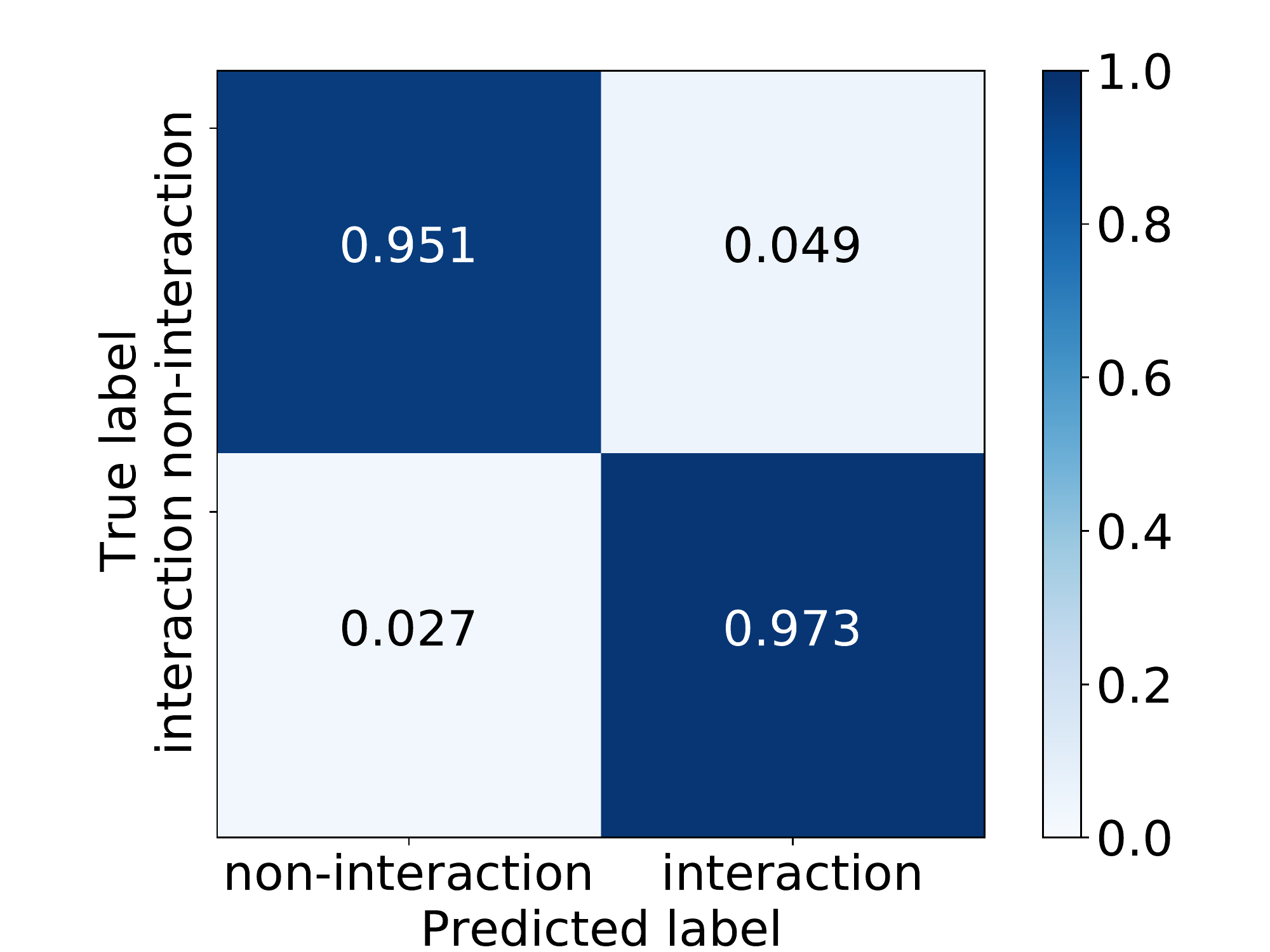}
    	\end{minipage}
    }
    \\
    \subfloat[NGY sliding window]{
    	\label{subfig:cm-NGY-sliding}
    	\hspace{-0.2cm}\begin{minipage}{0.24\textwidth}
    		\centering
    		\includegraphics[trim=0in 0in 0in 0in, width=\textwidth]{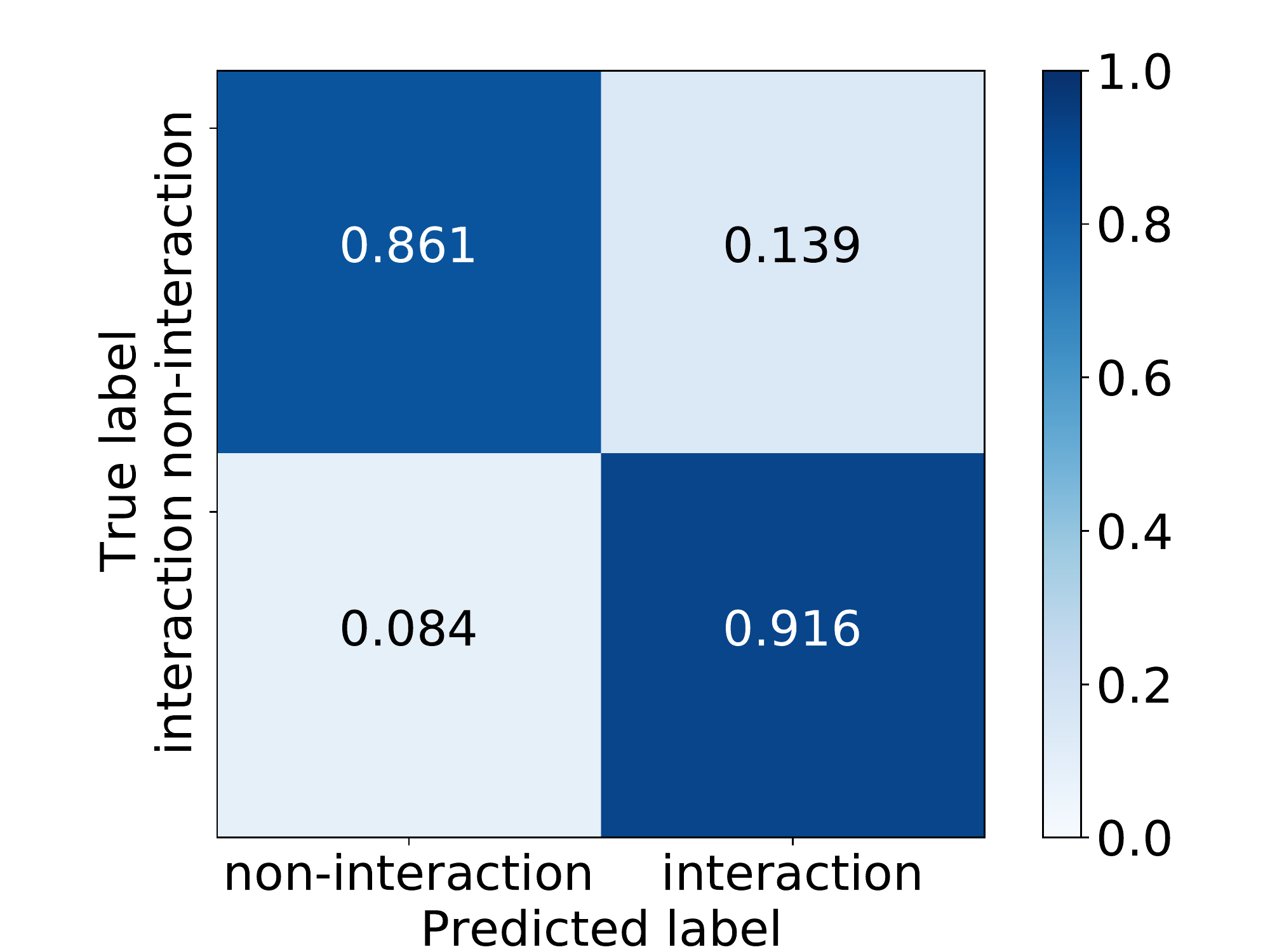}
    	\end{minipage}
    }
    \subfloat[NGY padding]{
    	\label{subfig:cm-NGY-pad}
    	\hspace{-0.2cm}\begin{minipage}{0.24\textwidth}
    		\centering
    		\includegraphics[trim=0in 0in 0in 0in, width=\textwidth]{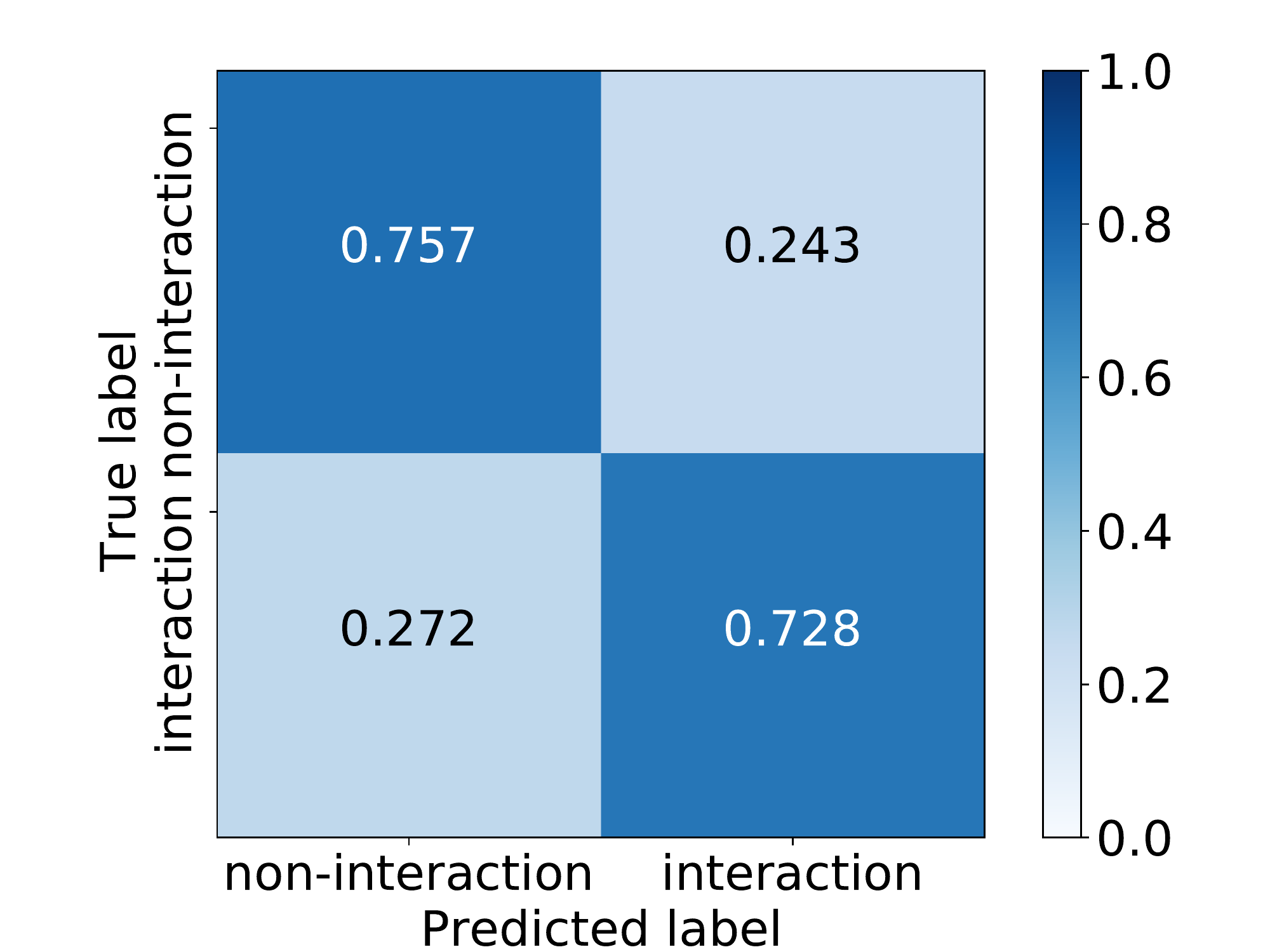}
    	\end{minipage}
    }
\caption[Confusion matrices of CVAE model]{Confusion matrices for the proposed CVAE model tested on the KoW/NGY dataset using the sliding window (a)/(c) and padding (b)/(d) methods. The confusion matrices are normalized so that they can be compared across sliding window and padding methods, as well as across the datasets.}  
\label{fig:cm-cvae}
\end{figure}

The confusion matrices for the proposed CVAE model using both the object and the optical--flow information are presented in Fig.~\ref{fig:cm-cvae}. It can be seen that the model using either the sliding window (true negative rate 0.970 and true positive rate 0.953) or the padding method (true negative rate 0.951 and true positive rate 0.973) achieves high performance for interaction detection tested on the KoW dataset. The proposed model achieves good performance using the sliding window method (true negative rate 0.861 and true positive rate 0.916) on the NGY dataset and maintains a relatively low false negative rate (0.084). Nevertheless, the performance of the proposed CVAE model tested on the NGY dataset is inferior to the one on the KoW dataset.
In contrast, the padding method only achieves mediocre performance (true negative rate 0.757 and true positive rate 0.728) tested on the NGY dataset. 

\subsection{Qualitative results}
\label{sec:InteDetcResults-qualitative}
The qualitative results intuitively showcase the process of interaction detection of the models. The fine--grained probability of the predicted interaction at each frame provides a clue of how the interaction intensity evolves over time. 

\begin{figure}[t!]
\centering
    \subfloat[Sliding window method.]{
    	\label{subfig:result-KoW-sliding}
    	\begin{minipage}{0.235\textwidth}
    		\centering
    		\includegraphics[trim=0in 0in 0in 0in, width=\textwidth]{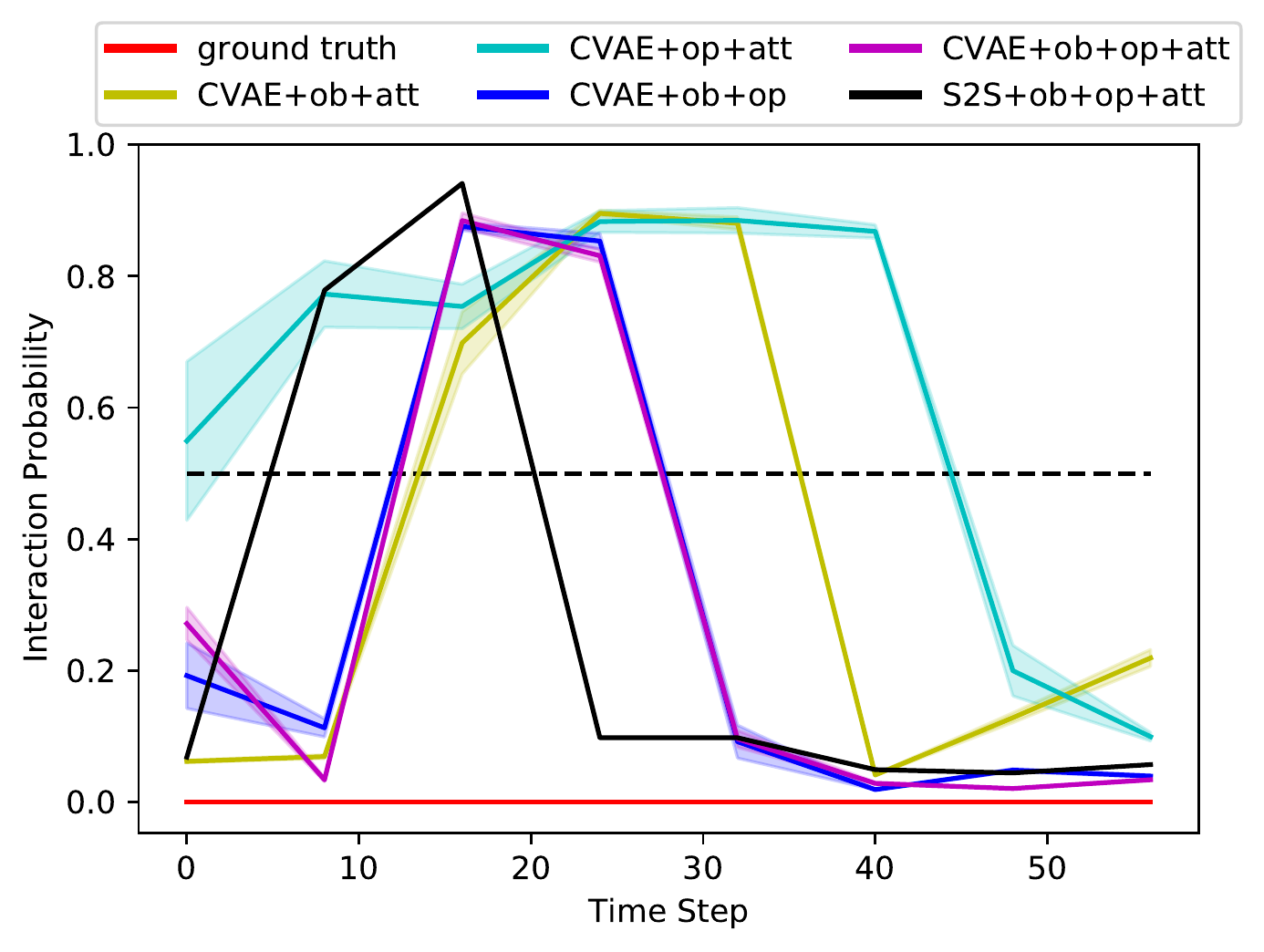}
    	\end{minipage}%
    }%
    \subfloat[Padding Method. ]{
    	\label{subfig:result-KoW-pad}
    	\begin{minipage}{0.235\textwidth}
    		\centering
    		\includegraphics[trim=0in 0in 0in 0in, width=\textwidth]{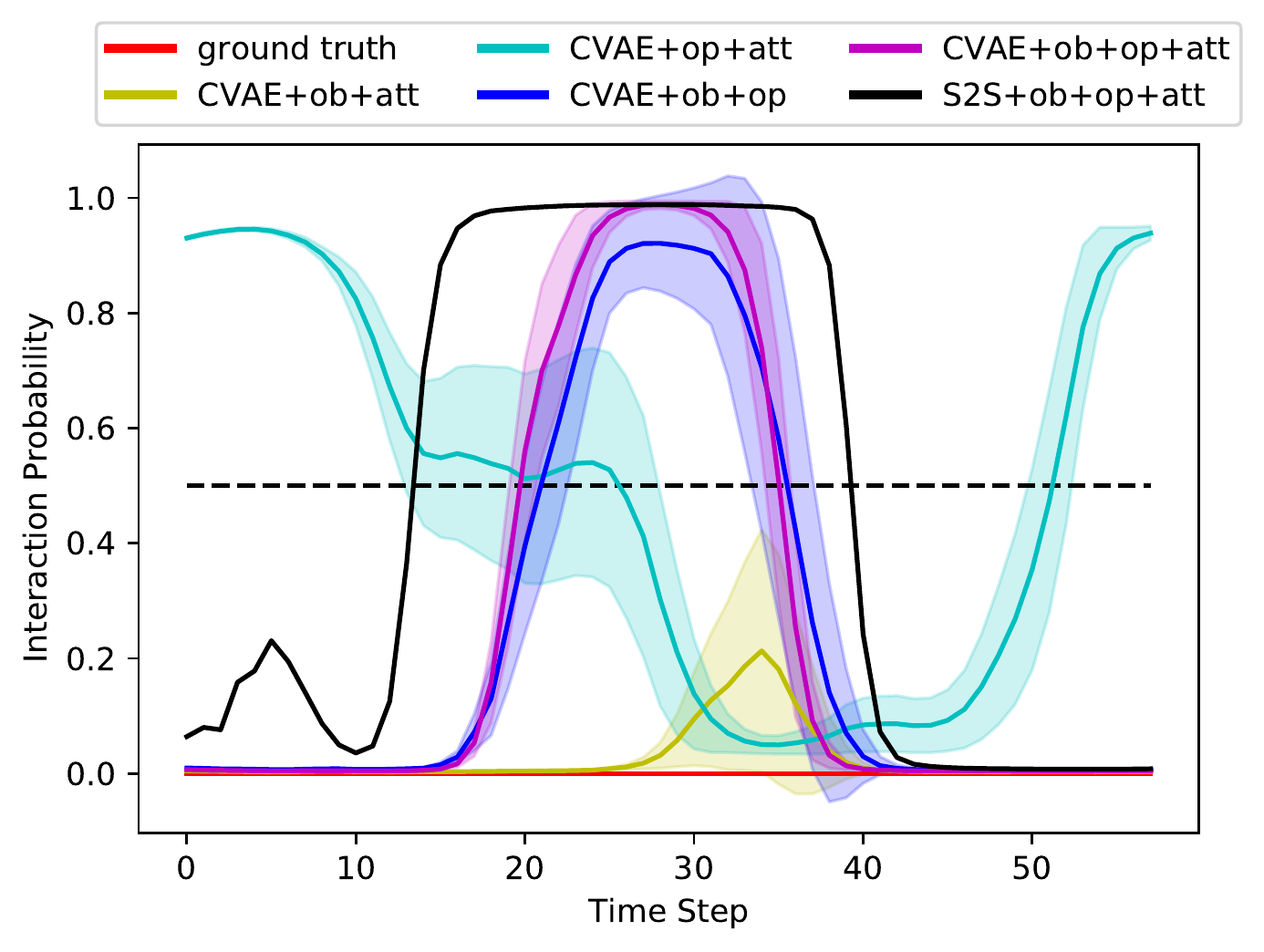}
    	\end{minipage}%
    }%
    \captionsetup[subfigure]{labelformat=empty}
     \subfloat[$\text{Time step}=0$]{
    	\label{subfig:result-KoW-sliding-00}
    	\begin{minipage}{0.23\textwidth}
    		\centering
    		\includegraphics[trim=0in 0in 0in 0in, width=\textwidth]{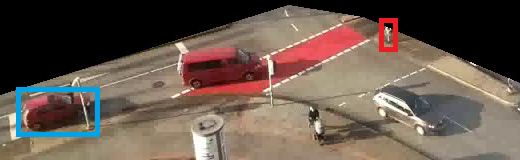}
    	\end{minipage}%
    }%
    \subfloat[$\text{Time step}=8$]{
    	\label{subfig:result-KoW-pad-08}
    	\begin{minipage}{0.23\textwidth}
    		\centering
    		\includegraphics[trim=0in 0in 0in 0in, width=\textwidth]{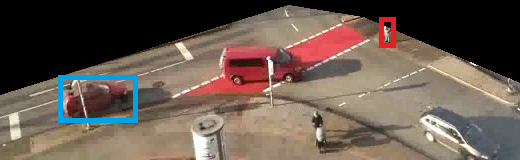}
    	\end{minipage}%
    }%
    
    \subfloat[$\text{Time step}=16$]{
    	\label{subfig:result-KoW-sliding-16}
    	\begin{minipage}{0.23\textwidth}
    		\centering
    		\includegraphics[trim=0in 0in 0in 0in, width=\textwidth]{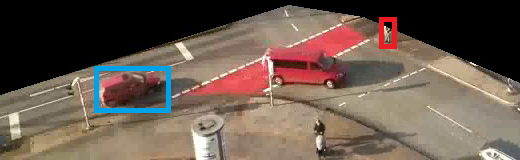}
    	\end{minipage}%
    }%
    \subfloat[$\text{Time step}=24$]{
    	\label{subfig:result-KoW-pad-24}
    	\begin{minipage}{0.23\textwidth}
    		\centering
    		\includegraphics[trim=0in 0in 0in 0in, width=\textwidth]{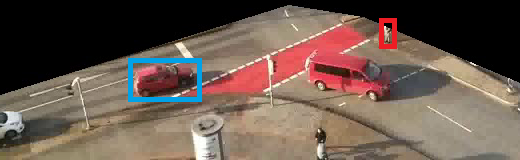}
    	\end{minipage}%
    }%
   
    \subfloat[$\text{Time step}=32$]{
    	\label{subfig:result-KoW-sliding-32}
    	\begin{minipage}{0.23\textwidth}
    		\centering
    		\includegraphics[trim=0in 0in 0in 0in, width=\textwidth]{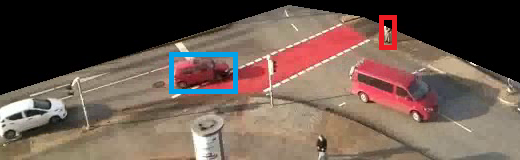}
    	\end{minipage}%
    }%
    \subfloat[$\text{Time step}=40$]{
    	\label{subfig:result-KoW-pad-40}
    	\begin{minipage}{0.23\textwidth}
    		\centering
    		\includegraphics[trim=0in 0in 0in 0in, width=\textwidth]{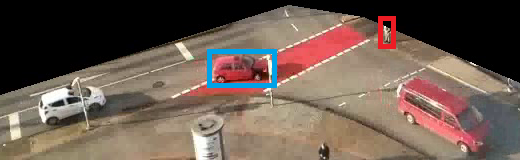}
    	\end{minipage}%
    }%
    
    \subfloat[$\text{Time step}=48$]{
    	\label{subfig:result-KoW-sliding-48}
    	\begin{minipage}{0.23\textwidth}
    		\centering
    		\includegraphics[trim=0in 0in 0in 0in, width=\textwidth]{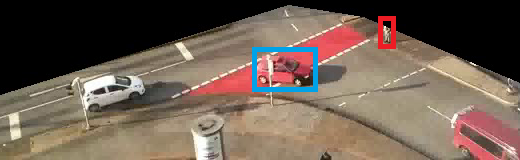}
    	\end{minipage}%
    }%
    \subfloat[$\text{Time step}=56$]{
    	\label{subfig:result-KoW-pad-56}
    	\begin{minipage}{0.23\textwidth}
    		\centering
    		\includegraphics[trim=0in 0in 0in 0in, width=\textwidth]{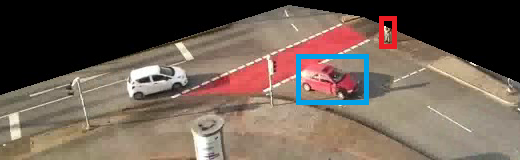}
    	\end{minipage}%
    }%
\caption[Examples of interaction probability for a right--hand intersection]{Examples of interaction probability at the frame level using the sliding window (a) and padding (b) methods, tested on the KoW dataset. The variance of the probabilities is visualized by the marginal shadow for the CVAE--based models. The corresponding video screenshots are aligned from upper left to the lower right at the bottom with a time interval of eight frames. The target vehicle is highlighted by the blue bounding box and the standstill pedestrian involved in the turning sequence is highlighted by the red bounding box.}  
\label{fig:results-KoW}
\end{figure}

Fig.~\ref{fig:results-KoW} demonstrates a non-interaction scenario at the KoW intersection between the right--turning target vehicle (in the blue bounding box) and the standstill pedestrian (in the red bounding box). There was no explicit interaction between them as the continuity of their behavior was not affected when the gap between them closed up, so the sequence was annotated as no interaction required. 
The sequence-level prediction is the average vote of all the frame-level predictions. 
At the sequence level, all the models correctly predict this scenario as non-interaction using both the sliding window (Fig.~\ref{subfig:result-KoW-sliding}) and padding (Fig.~\ref{subfig:result-KoW-pad}) methods, except the ablative model \textit{[CVAE+op+att]} (in cyan) that only uses the optical--flow information. However, all the models predict a high probability of interaction when the vehicle approached the pedestrian. 
Also, the variance of the CVAE--based models in Fig.\ref{subfig:result-KoW-pad} increases when the probabilities change from under 0.5 to a higher value of interaction.
The baseline model \textit{[S2S+ob+op+att]} (in black) generates a similar pattern in frame--wise predictions. But it is deterministic at each frame and does not have the mechanism to represent the uncertainty of the predictions. 

\begin{figure}[t!]
\centering
    \subfloat[Sliding window method.]{
    	\label{subfig:result-NGY-sliding}
    	\begin{minipage}{0.235\textwidth}
    		\centering
    		\includegraphics[trim=0in 0in 0in 0in, width=\textwidth]{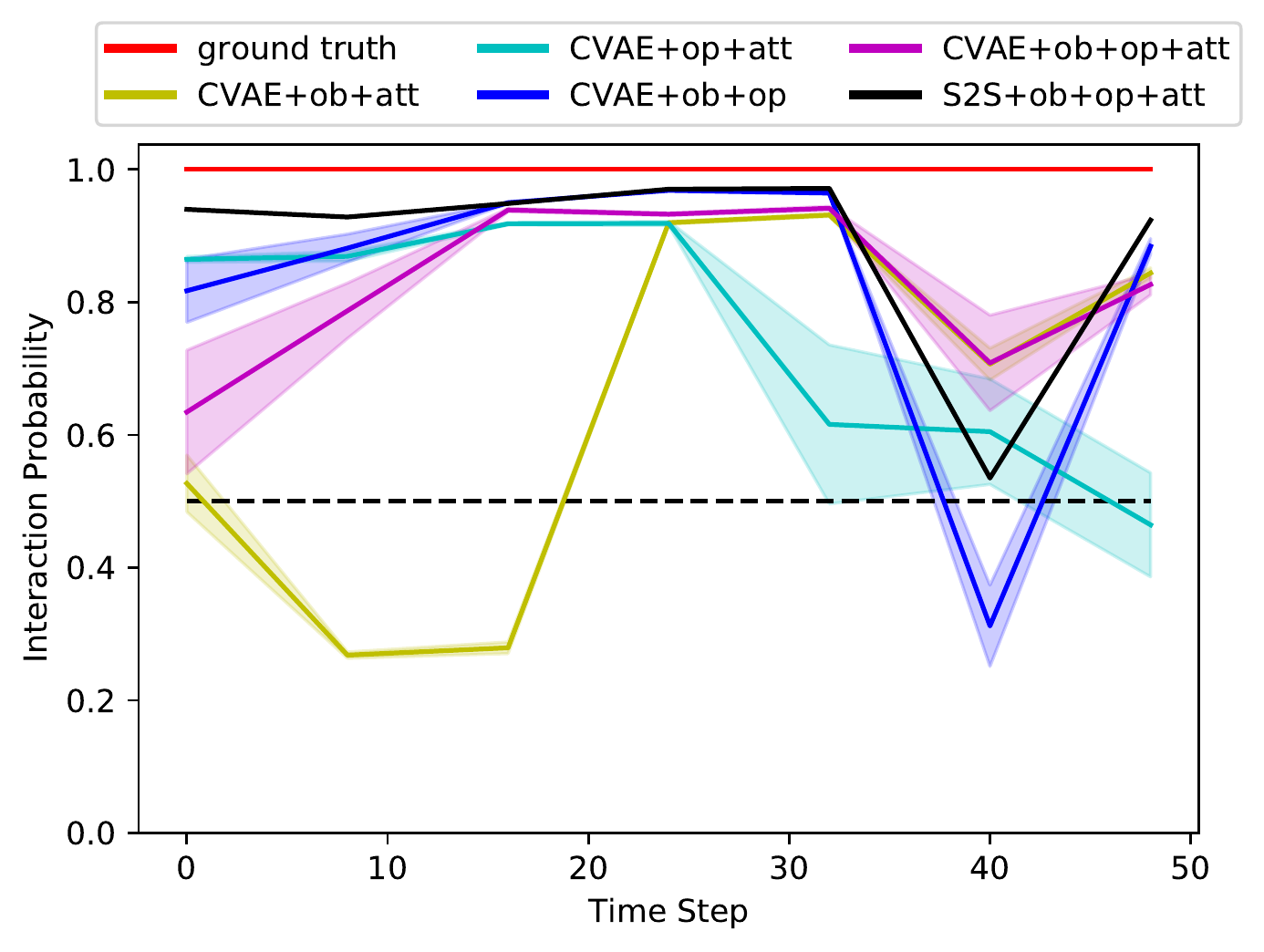}
    	\end{minipage}%
    }%
    \subfloat[Padding Method.]{
    	\label{subfig:result-NGY-pad}
    	\begin{minipage}{0.235\textwidth}
    		\centering
    		\includegraphics[trim=0in 0in 0in 0in, width=\textwidth]{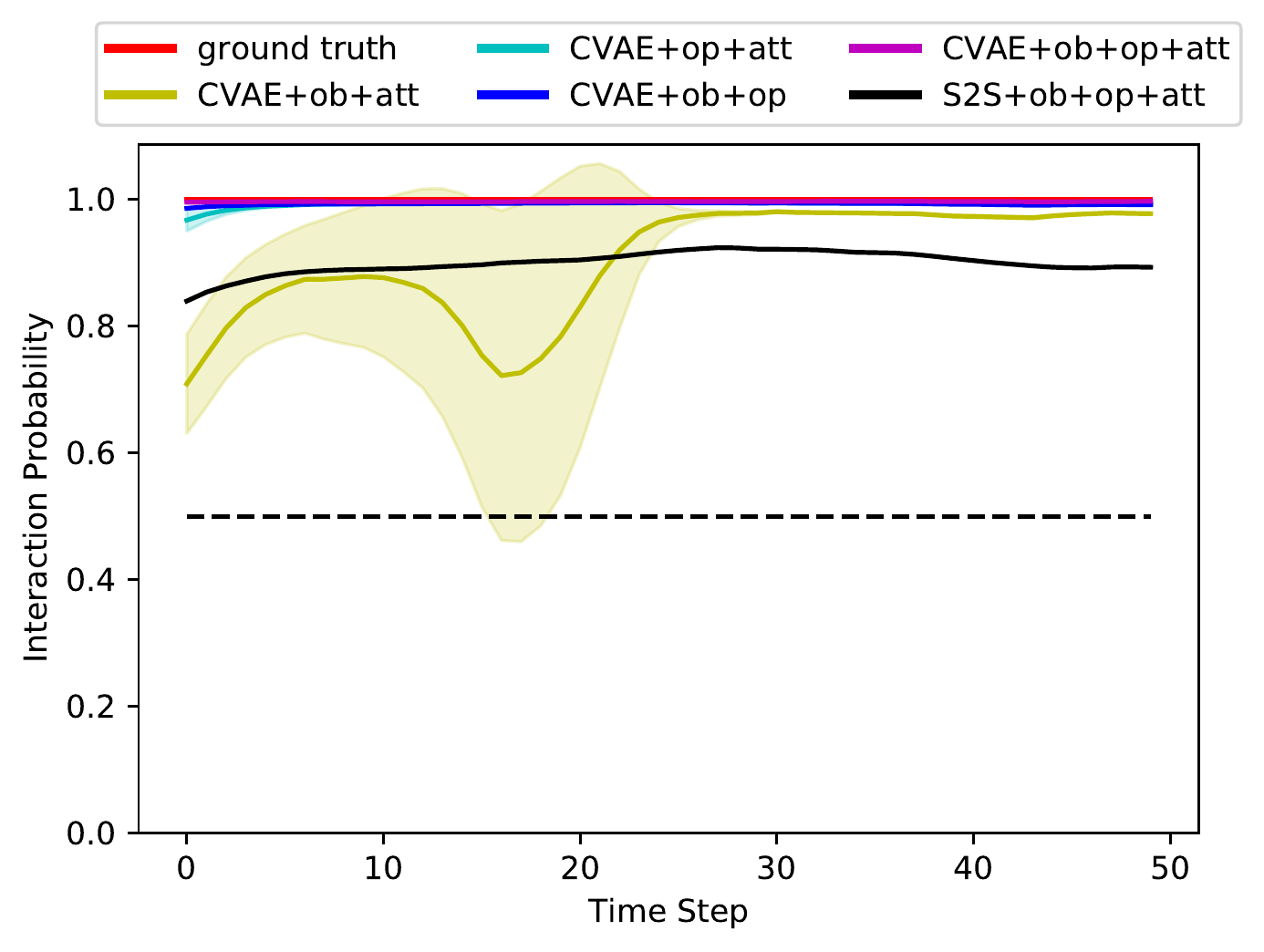}
    	\end{minipage}%
    }%
    \captionsetup[subfigure]{labelformat=empty}
    \subfloat[$\text{Time step}=0$]{
    	\label{subfig:result-NGY-sliding-300}
    	\begin{minipage}{0.23\textwidth}
    		\centering
    		\includegraphics[trim=0in 1.3in 0in 0in, clip=true, width=\textwidth]{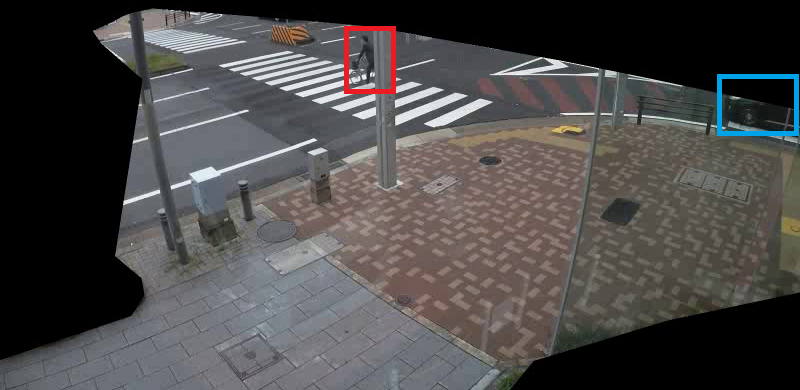}
    	\end{minipage}%
    }%
    \subfloat[$\text{Time step}=8$]{
    	\label{subfig:result-NGY-pad-308}
    	\begin{minipage}{0.23\textwidth}
    		\centering
    		\includegraphics[trim=0in 1.3in 0in 0in, clip=true, width=\textwidth]{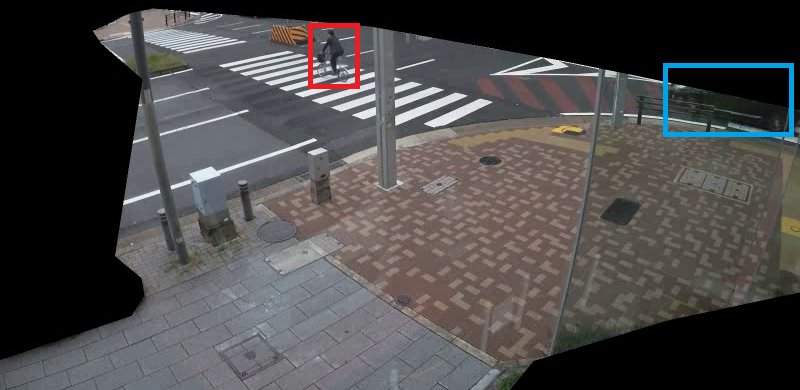}
    	\end{minipage}%
    }%
    
    \subfloat[$\text{Time step}=16$]{
    	\label{subfig:result-NGY-sliding-316}
    	\begin{minipage}{0.23\textwidth}
    		\centering
    		\includegraphics[trim=0in 1.3in 0in 0in, clip=true, width=\textwidth]{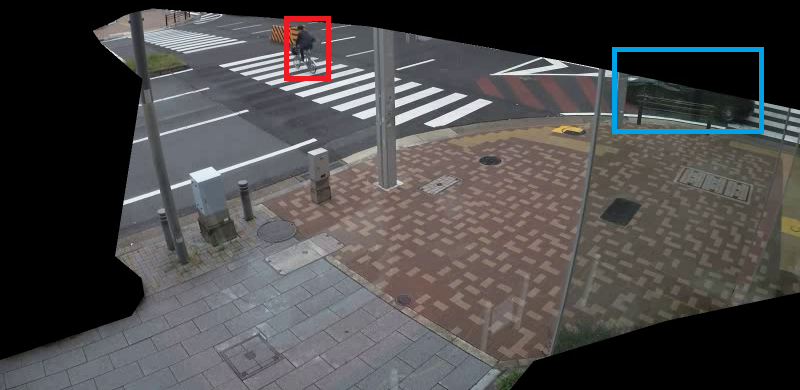}
    	\end{minipage}%
    }%
    \subfloat[$\text{Time step}=24$]{
    	\label{subfig:result-NGY-pad-324}
    	\begin{minipage}{0.23\textwidth}
    		\centering
    		\includegraphics[trim=0in 1.3in 0in 0in, clip=true, width=\textwidth]{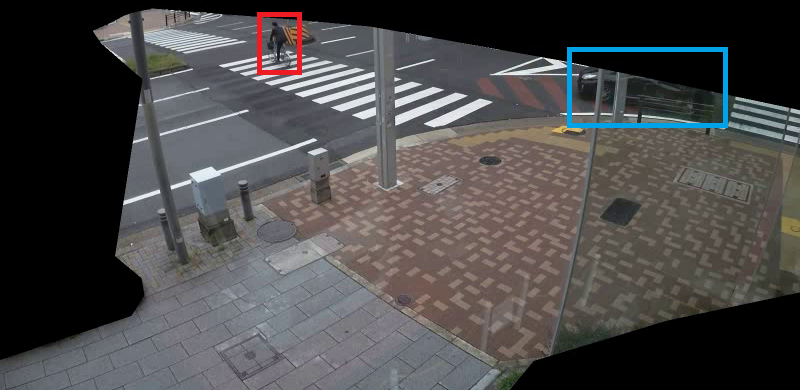}
    	\end{minipage}%
    }%
    
    \subfloat[$\text{Time step}=32$]{
    	\label{subfig:result-NGY-sliding-332}
    	\begin{minipage}{0.23\textwidth}
    		\centering
    		\includegraphics[trim=0in 1.3in 0in 0in, clip=true, width=\textwidth]{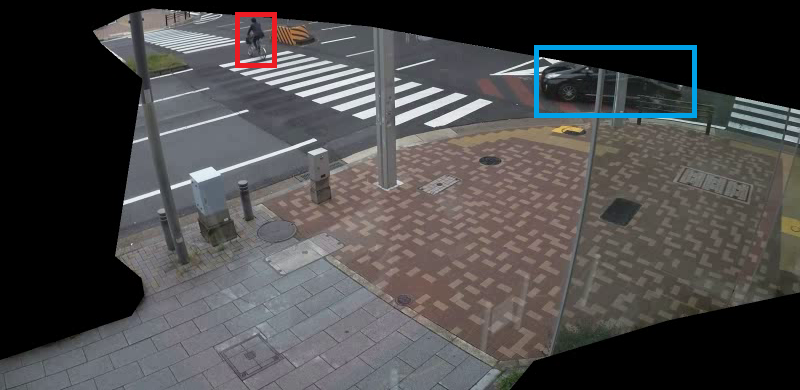}
    	\end{minipage}%
    }%
    \subfloat[$\text{Time step}=40$]{
    	\label{subfig:result-NGY-pad-340}
    	\begin{minipage}{0.23\textwidth}
    		\centering
    		\includegraphics[trim=0in 1.3in 0in 0in, clip=true, width=\textwidth]{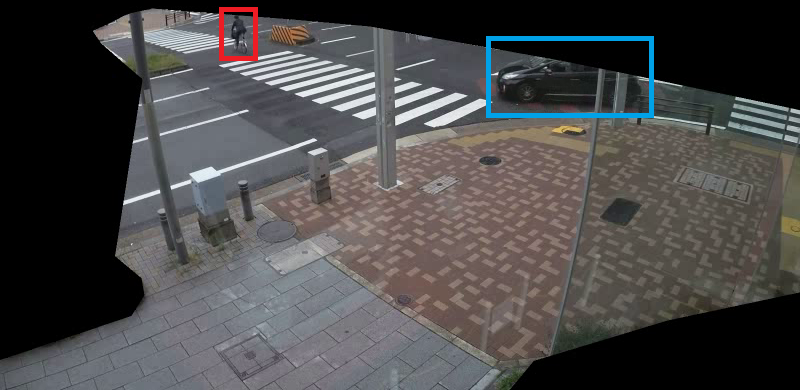}
    	\end{minipage}%
    }%
    
    \subfloat[$\text{Time step}=48$]{
    	\label{subfig:result-NGY-sliding-348}
    	\begin{minipage}{0.23\textwidth}
    		\centering
    		\includegraphics[trim=0in 1.3in 0in 0in, clip=true, width=\textwidth]{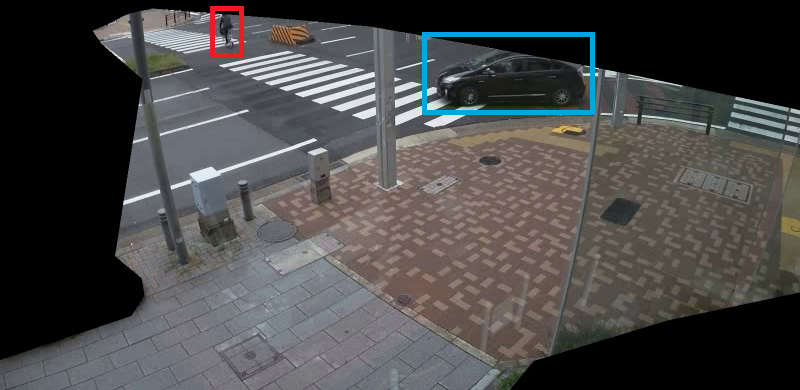}
    	\end{minipage}
    }%
\caption[Examples of interaction probability for a left--hand intersection]{Examples of interaction probability at the frame level using the sliding window (a) and padding (b) methods, tested on the NGY dataset. The variance of the probabilities is visualized by the marginal shadow for the CVAE--based models. The corresponding video screenshots are aligned from upper left to the lower middle at the bottom with a time interval of eight frames. The target vehicle is highlighted by the blue bounding box and the passing cyclist involved in the turning sequence is highlighted by the red bounding box.}  
\label{fig:results-NGY}
\end{figure}

Fig.~\ref{fig:results-NGY} demonstrates an interaction scenario at the NGY intersection between the left--turning target vehicle (in the blue bounding box) and the crossing cyclist (in the red bounding box). Interaction was required between them as the vehicle had to decelerate or even briefly stop, yielding the way to the cyclist. All the models correctly predict this sequence as interaction using both the sliding window (Fig.~\ref{subfig:result-NGY-sliding}) and padding (Fig.~\ref{subfig:result-NGY-pad}) methods. Similar to the scenario above, the variance of the probabilities for the CVAE--based models using the sliding window method change with the modification of the distance between the target vehicle and the cyclist. As the distance between them decreases, the probability is higher and the variance is smaller for interaction, and vice versa. 
The ablative model based on the object information using the padding method has higher uncertainty levels than the other models of the frame--wise predictions.   

\subsection{Analysis of the results}
\label{subsec:interactionresultsanalysis}
The results shown above are analyzed with respect to: (\RNum{1}) the pros and cons between the sliding window and padding methods; (\RNum{2}) the performance between the proposed CVAE model and the baseline model; (\RNum{3}) the contribution of the object information and the optical--flow information via the ablative models; (\RNum{4}) the impact of the self-attention mechanism.

The performances of the sliding window and padding methods are not only influenced by the size of the training data, but also the zero--padded values. The sliding window method does not depend on the sequence length, which is more flexible in dealing with various sequence lengths. Hence, the number of training samples was not compromised for the experiments. On the other hand, the padding method requires a pre-defined fixed sequence length, which is unable to deal with longer sequences. Hence, the number of training samples was compromised by excluding longer sequences. The impact of the training data size has been shown by the performance difference across the KoW and NGY datasets. The numbers of the training samples of KoW for the sliding window and padding methods are similar (see Table~\ref{tb:interactiondatapartition}), and their performances for interaction detection are comparable to each other (see Table~\ref{tb:KoW-results}). On the contrary, the number of training samples of NGY for the sliding window method is much larger than the one for the padding method (see Table~\ref{tb:interactiondatapartition}). The prediction by  the sliding window method is more accurate than the padding method (see Table~\ref{tb:NGY-results}). In addition, the shorter sequences were padded with zeros. This is problematic for the information extracted by optical flow. Because the zero values in the optical--flow feature vector represent the background of the intersection or static road users. Even though a padding mask is incorporated into the sequence for indicating the actual sequence length, the negative impact cannot be fully remedied due to the complex learning process in training. The negative impact of padded zeros from the padding method has been uncovered by the impaired performance of the ablative model \textit{[CVAE+op+att]} compared to the sliding window method. 

In general, the proposed CVAE model \textit{[CVAE+ob+op+att]} outperforms the baseline model [\textit{S2S+ob+op+att}] quantitatively (see Table~\ref{tb:KoW-results} and \ref{tb:NGY-results}) and qualitatively (see Fig.~\ref{fig:results-KoW} and \ref{fig:results-NGY}). In the CVAE model, the latent variables $\mathbf{z}$ are trained to capture the stochastic attributes of road users' behavior in various traffic situations, which is optimized by the Kullback-Leibler divergence loss against a Gaussian prior. In addition to the Kullback-Leibler divergence loss, the reconstruction loss is trained by minimizing the cross-entropy loss between ground truth and prediction.
Optimizing these two losses together enables the CVAE model to generate diverse predictions. With the multi--sampling process of the latent variables, the predicted probabilities of interaction at each frame vary, especially when the probabilities change over time, see Fig.~\ref{fig:results-KoW} and \ref{fig:results-NGY}; the variance of the probabilities indicates the uncertainty in the predictions. In contrast, the baseline model is trained only by optimizing the reconstruction loss. It tends to learn the ``average'' behavior of road users. Predictions by the baseline model are rather deterministic and there is no mechanism to interpret the uncertainty of the predictions.

The combined information of object detection and optical flow shows a stable performance for the interaction detection task. The performance of interaction detection highly depends on the quality of the input information extracted from videos, which is often impaired for many reasons. A single type of information may not be sufficient for this task.
As indicated by the limited performance of the ablative models that only use optical--flow information on the KoW dataset, without the object information the noisy optical--flow information from the through lane or padded zeros may impact the detection performance. Similarly, the ablative models that only use object information on the NGY dataset also achieved limited performance. The distorted object information, especially for the road users close to the camera at the NGY dataset, could lead to wrong interaction detection.
The combination of the extraction techniques increases the possibility to maintain a good quality of the input information, so as to achieve a stable performance of interaction detection.

The self-attention mechanism does not show a consistent benefit across the datasets. The CVAE models regardless of the self-attention mechanism yield very similar results for interaction detection using both the sliding window and padding methods on the KoW dataset, and using the padding method on the NGY dataset. The improvement with the self-attention mechanism can be found for the sliding window method on the NGY dataset. First, the self-attention layer is followed by an LSTM (Fig.~\ref{fig:pipelineInteractionDetection}), which may be redundant for learning the interconnections along the time axis. The self-attention layer is likely under-trained due to the small dataset size or redundant layers, whereas the LSTM is already sufficient for learning the temporal patterns of the sequence data from the KoW intersection. On the other hand, the sequence data from the NGY intersection is more complex, \eg~longer and more varying sequence lengths (see Fig.~\ref{fig:seqdistrubitions}) and high traffic density. On top of the LSTM, the self-attention mechanism has turned out to be beneficial for further learning the temporal patterns.

In summary, the sliding window method is more flexible than the padding method in dealing with various sequence lengths. The CVAE models using the combined information of both object detection and optical flow achieve a more stable performance compared to using a single type of information. The multi--sampling process enables the CVAE--based models to mimic the uncertainty of road users' behavior, and the self--attention mechanism is only beneficial for learning temporal patterns from complex data. Overall the proposed model \textit{[CVAE+ob+op+att]} using the sliding window method achieves a more desirable performance across the datasets.

\section{Discussion}
\label{sec:InteDetcDiscussion}
Here we discuss the failed detection by the proposed CVAE model using the sliding window method and the challenges to transfer the model from one intersection to the other smoothly.

\subsection{Failed detection}
\label{subsec:faileddetection}

\begin{figure}[t!]
\captionsetup[subfigure]{labelformat=empty}
\centering
    \subfloat[FN-\RNum{1}]{
    	\label{subfig:KoW-FN-a}
    	\begin{minipage}{0.23\textwidth}
    		\centering
    		\includegraphics[trim=0in 0in 0in 0in, width=\textwidth]{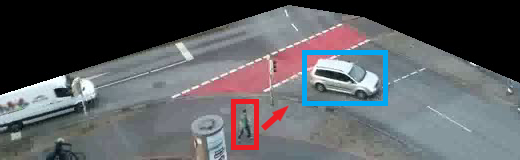}
    	\end{minipage}
    }
    \subfloat[FN-\RNum{2}]{
    	\label{subfig:KoW-FN-b}
    	\begin{minipage}{0.23\textwidth}
    		\centering
    		\includegraphics[trim=0in 0in 0in 0in, width=\textwidth]{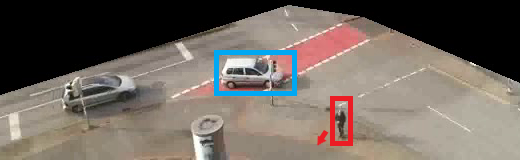}
    	\end{minipage}
    }
    \caption[Examples of false negative detection on the KoW dataset]{Examples of the false negative detection on the KoW dataset. The right--turning target vehicles are denoted by the blue bounding boxes and the involved VRUs are denoted by the red bounding boxes.}  
\label{fig:KoW-falsenegatives}
\end{figure}

\begin{figure}[t!]
\captionsetup[subfigure]{labelformat=empty}
\centering
    \subfloat[FN-\RNum{1}]{
    	\label{subfig:NGY-FN-a}
    	\begin{minipage}{0.23\textwidth}
    		\centering
    		\includegraphics[trim=0in 1.3in 0in 0in, clip=true, width=\textwidth]{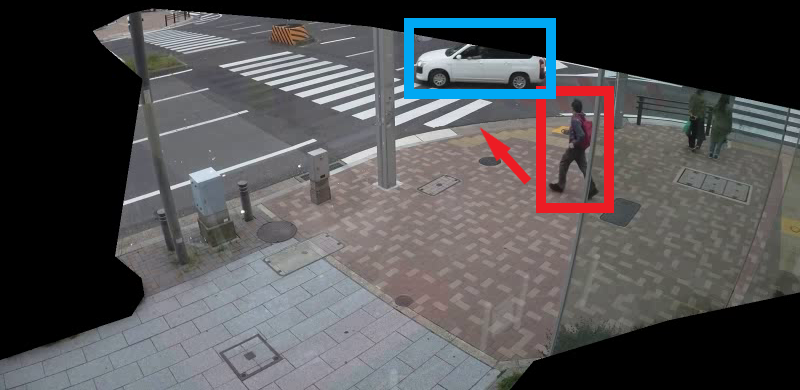}
    	\end{minipage}%
    }%
    \subfloat[FN-\RNum{2}]{
    	\label{subfig:NGY-FN-b}
    	\begin{minipage}{0.23\textwidth}
    		\centering
    		\includegraphics[trim=0in 1.3in 0in 0in, clip=true, width=\textwidth]{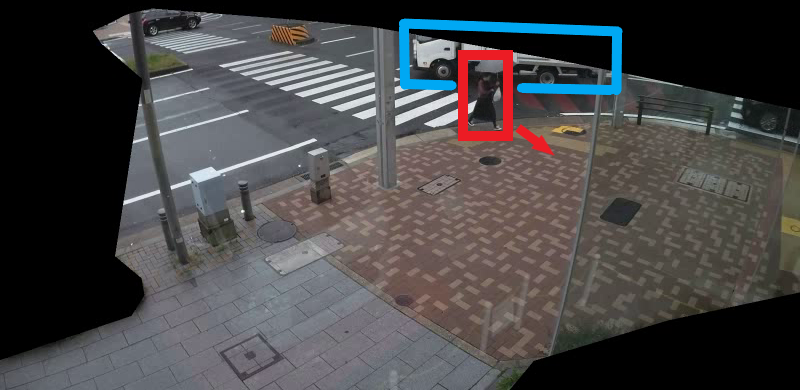}
    	\end{minipage}%
    }%

    \subfloat[FN-\RNum{3}]{
    	\label{subfig:NGY-FN-c}
    	\begin{minipage}{0.23\textwidth}
    		\includegraphics[trim=0in 1.3in 0in 0in, clip=true, width=\textwidth]{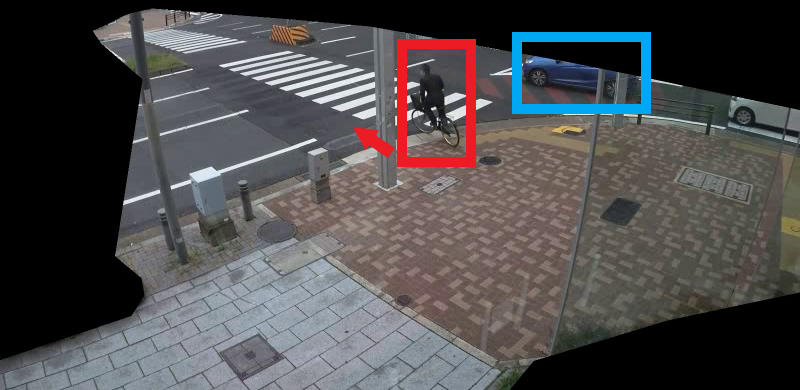}
    	\end{minipage}%
    }%
    \caption[Examples of false negative detection on the NGY dataset]{Examples of the false negative detection on the NGY dataset. The left--turning target vehicles are denoted by the blue bounding boxes and the involved VRUs are denoted by the red bounding boxes.}  
\label{fig:NGY-falsenegatives}
\end{figure}

\begin{figure}[t!]
\captionsetup[subfigure]{labelformat=empty}
\centering
    \subfloat[FP-\RNum{1}]{
    	\label{subfig:KoW-FP-a}
    	\begin{minipage}{0.23\textwidth}
    		\centering
    		\includegraphics[trim=0in 0in 0in 0in, width=\textwidth]{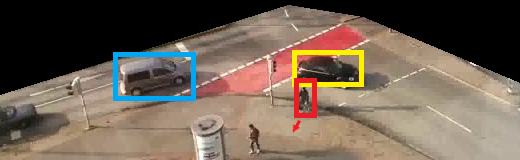}
    	\end{minipage}
    }
    \subfloat[FP-\RNum{4}]{
    	\label{subfig:KoW-FP-d}
    	\begin{minipage}{0.23\textwidth}
    		\centering
    		\includegraphics[trim=0in 0in 0in 0in, width=\textwidth]{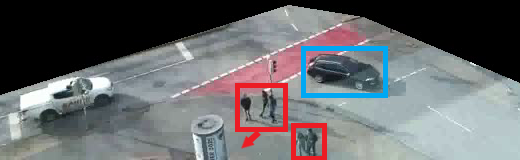}
    	\end{minipage}
    }
\caption[Examples of the false positive detection on the KoW dataset]{Examples of the false positive detection on the KoW dataset. The right--turning target vehicles are denoted by the blue bounding boxes, the leading, but not target vehicle, is denoted by the yellow bounding box, and the involved VRUs are denoted by the red bounding boxes.}  
\label{fig:KoW-falsepositives}
\end{figure}

\begin{figure}[t!]
\captionsetup[subfigure]{labelformat=empty}
\centering
    \subfloat[FP-\RNum{1}]{
    	\label{subfig:NGY-FP-a}
    	\begin{minipage}{0.23\textwidth}
    		\centering
    		\includegraphics[trim=0in 1.1in 0in 0in, clip=true, width=\textwidth]{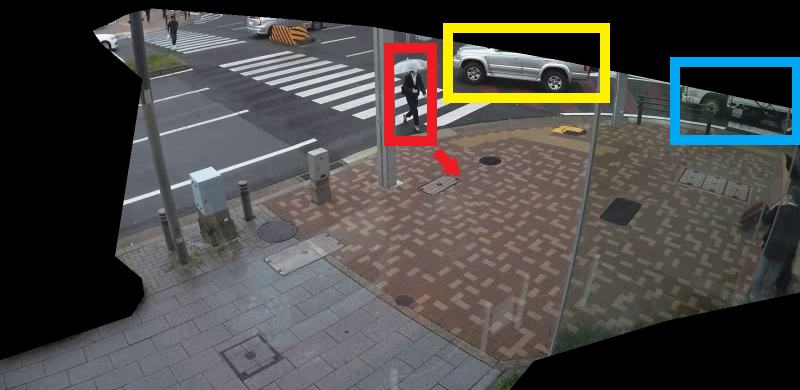}
    	\end{minipage}%
    }%
    \subfloat[FP-\RNum{2}]{
    	\label{subfig:NGY-FP-b}
    	\begin{minipage}{0.23\textwidth}
    		\centering
    		\includegraphics[trim=0in 1.1in 0in 0in, clip=true, width=\textwidth]{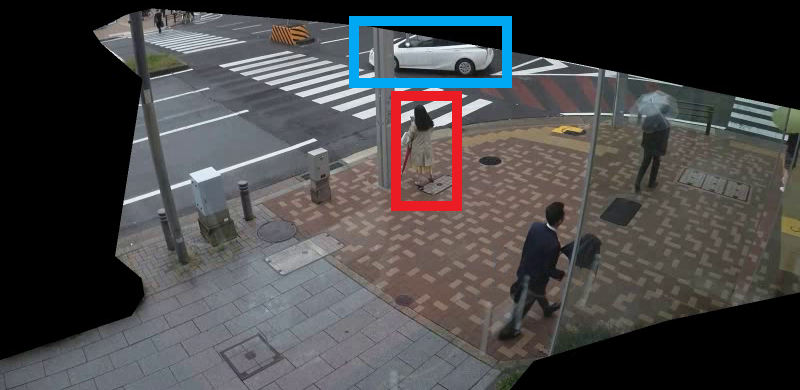}
    	\end{minipage}%
    }%
    
    \subfloat[FP-\RNum{3}]{
    	\label{subfig:NGY-FP-c}
    	\begin{minipage}{0.23\textwidth}
    		\centering
    		\includegraphics[trim=0in 1.1in 0in 0in, clip=true, width=\textwidth]{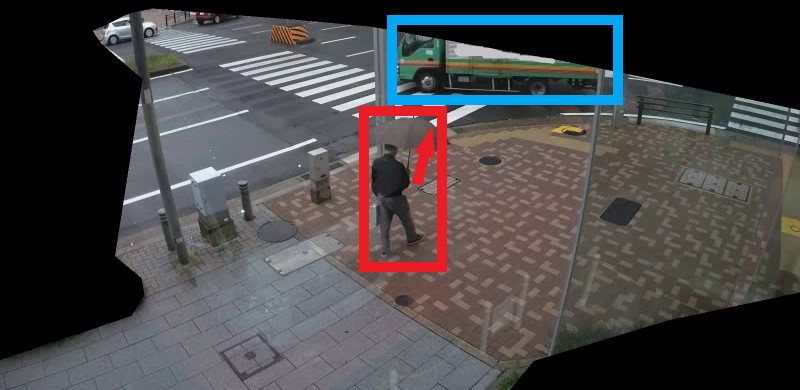}
    	\end{minipage}%
    }%
    \subfloat[FP-\RNum{4}]{
    	\label{subfig:NGY-FP-d}
    	\begin{minipage}{0.23\textwidth}
    		\centering
    		\includegraphics[trim=0in 1.1in 0in 0in, clip=true, width=\textwidth]{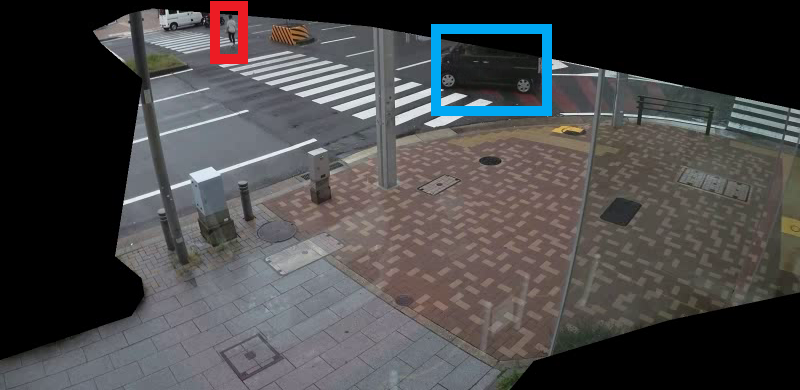}
    	\end{minipage}%
    }%
\caption[Examples of the false positive on the NGY dataset]{Examples of the false positive on the NGY dataset. The left--turning target vehicles are denoted by the blue bounding boxes, the leading, but not target vehicle, is denoted by the yellow bounding box, and the involved VRUs are denoted by the red bounding boxes.}  
\label{fig:NGY-falsepositives}
\end{figure}

Various reasons can lead to a wrong interaction classification. Table~\ref{tb:intersection-wrongdetection} categorizes the wrongly detected scenarios \ie~false negative (FN) and false positive (FP) tested on both the KoW and NGY datasets.
The false negative examples are visualized in Fig.~\ref{fig:KoW-falsenegatives} for KoW and in Fig.~\ref{fig:NGY-falsenegatives} for NGY, and the false positive examples are visualized in Fig.~\ref{fig:KoW-falsepositives} for KoW and in Fig.~\ref{fig:NGY-falsepositives} for NGY.
\begin{table}[hbpt!]
\caption[Categories of wrongly detected scenarios]{Categories of the wrongly detected scenarios by \textit{[CVAE+ob+op+att]} using the sliding window method.}
\setlength{\tabcolsep}{2.5pt}
\centering
\begin{tabular}{lllll}
\\ \toprule
Errors            & Scenario description                                          & Category      & KoW & NGY \\ \hhline{=====}
                  & pedestrian entering the intersection                          & (FN-\RNum{1}) & 8   & 7   \\
FN                & pedestrian leaving the intersection                           & (FN-\RNum{2}) & 1   & 4   \\
                  & cyclist entering the intersection                             & (FN-\RNum{3}) & -   & 2   \\ \midrule
                  & car following                                                 & (FP-\RNum{1}) & 4   & 17  \\
FP                & pedestrian standing near the intersection & (FP-\RNum{2}) & -   & 3   \\
                  & pedestrian approaching from the sidewalk                      & (FP-\RNum{3}) & -   & 1   \\
                  & pedestrian finishing crossing                                 & (FP-\RNum{4}) & 1   & 1   \\ \midrule
Total$^*$         &                                                               &               & 14  & 35  \\ \bottomrule
\end{tabular}
\label{tb:intersection-wrongdetection}
\begin{tabular}{@{}c@{}}
\multicolumn{1}{p{3.3in}}{$^{*}$The total number of the wrongly detected scenarios listed here is slightly different as shown in the above confusion matrices due to the multi--sampling of the CVAE model.}
\end{tabular}
\end{table}

The FN scenarios are associated with VRUs entering (FN-\RNum{1} and FN-\RNum{3}) or leaving (FN-\RNum{2}) the intersection space. Due to their relatively long distance to the target vehicle, but fast travel speed, they are erroneously classified as non-interaction. 

Most of the FP scenarios are associated with the target vehicle following a leading vehicle. As exemplified by FP-\RNum{1} in Fig.~\ref{fig:KoW-falsepositives}~and~\ref{fig:NGY-falsepositives}, only the leading vehicle (in the yellow bounding box) required direct interactions with the involved VRUs. After the leading vehicle finished turning, the pedestrian (in the red bounding box) also completed crossing. Afterwards, there was no interaction required from the target vehicle (in the blue bounding box) with the VRUs. However, the CVAE model has limited performance in handling this type of situations. Because the current model does not have explicit information to differentiate the leading and target vehicles, and the model is not specifically trained for car following situations.

In addition, a short distance from the VRUs to the intersection, \eg~standing on the sidewalk close to the intersection (FP-\RNum{2} Fig.~\ref{fig:NGY-falsepositives}) or just finishing crossing (FP-\RNum{4}, Fig.~\ref{fig:KoW-falsepositives}~and~\ref{fig:NGY-falsepositives}), can also lead to an FP case. 
The distortion of the distance may lead to an FP case as well. For example, in FP-\RNum{3} in Fig.~\ref{fig:NGY-falsepositives}, even though the pedestrian on the sidewalk was relatively far from the turning vehicle, it has been still classified as an interaction by the model due to the distorted distance at the NGY intersection. However, the camera at the KoW intersection was installed at a higher elevation than the camera at the NGY intersection. The distortion is thus less harmful for the horizontal distance. Among other reasons, this might have contributed to the better performance of the model tested on the KoW dataset than on the NGY dataset. 

Based on the discussion of the failed detection scenarios, the limitations of the proposed model are summarized as follows: 
    1) The definition of interaction only considers the relationship between the target turning vehicle and the involved VRUs. The car--following relationship is not included. Without considering this relationship often leads to false positive detection between the following car and the VRUs.
    2) The crossing directions of VRUs are not used as a factor to differentiate interaction types. For example, interactions between a turning vehicle and pedestrians or cyclists that are approaching the crossing area from near side and far side are labeled as the same interaction type. However, the discussion above indicates that the moving directions of VRUs are important for estimating the interactions between the turning vehicle and VRUs, especially when the VRUs are leaving the intersection.
    3) The exact distances between a turning vehicle and the involved VRUs are not measured, thus the change of the distances between them cannot be correctly quantified. Without the measurement of distance, it is difficult for a model to distinguish the subtle difference between interaction and non-interaction. Especially, when the image distance is distorted by the camera's perspective or when an occlusion happens, the model's performance will be impaired. 

\subsection{Challenges of cross dataset generalization}
\label{subsec:crossdatavalidation}

The models proposed in this chapter are adapted for interaction detection at different intersections, in order to analyze the generalizability of the above models.
In the previous experiment setting, all the models were trained and tested using the dataset from the same intersections. In this section, the models trained using the KoW dataset were tested on the NGY dataset, and vice versa. Frames from the test set were resized into the same size and mirrored into the same direction as the training set, so that the trained models could be tested on both datasets without changing the setting of the input size.

Table~\ref{tb:crossvalidation-results} lists the results for the cross dataset validation.  It can be seen that, both the CVAE--based and sequence-to-sequence encoder--decoder models do not achieve good performance either using the sliding window or padding method. This could be because these two datasets (Sec.~\ref{sec:InteDetcData}) are very different in terms of, \eg~vehicle's travel direction, camera's perspective, frame size and rate, sequence length, intersection layout, traffic density, and cultural factors (Germany vs. Japan).  Note that, because the camera parameters and reference coordinates were not available from the datasets,
in this paper projection is not applied to transform the perspective to a bird's-eye view. Under the cross dataset validation setting, the resized frames distort the motion and position of the dynamic objects and confused the models for predicting the interactions between vehicles and VRUs. However, this leads to the future research question---how to generalize the models for different intersections and traffic, and even different cultures?

\begin{table}[hbpt!]
\caption[Performance of cross dataset validation]{Performance of cross dataset validation for interaction detection on the KoW and NGY datasets.}
\centering
\setlength{\tabcolsep}{1.4pt}
\begin{tabular}{llllll}
\toprule

Model                     & Shape    & Accuracy                  & Precision                 & Recall                    & F1-score                       \\ \hhline{======}
\multicolumn{6}{c}{{Trained on the NGY dataset and tested on the KoW dataset}}  \\ \midrule
\textit{[S+ob+op+att]}  & sli.  & 0.490                     & 0.430                     & 0.490                     & 0.350                          \\ 
\textit{[C+ob+op+att]} & sli.  & 0.490{$_{\pm0.003}$}            & 0.495{$_{\pm0.001}$}          & 0.934{$_{\pm0.005}$}          & 0.647{$_{\pm0.002}$}               \\  
\textit{[S+ob+op+att]}  & pad.     & 0.473                     & 0.450                     & 0.470                     & 0.400                          \\ 
\textit{[C+ob+op+att]} & pad.      & 0.485{$_{\pm0.002}$}            & 0.491{$_{\pm0.001}$}          & 0.804{$_{\pm0.004}$}          & 0.609{$_{\pm0.001}$}               \\ \hhline{======}
\multicolumn{6}{c}{{Trained on the KoW dataset and tested on the NGY dataset}}  \\ \midrule
\textit{[S+ob+op+att]}  & sli.  & 0.535                     & 0.526                     & 0.704                     & 0.602                          \\ 
\textit{[C+ob+op+att]} & sli.  & 0.541{$_{\pm0.005}$}            & 0.540{$_{\pm0.005}$}          & 0.557{$_{\pm0.007}$}          & 0.548{$_{\pm0.006}$}      \\ 
\textit{[S+ob+op+att]}  & pad.      & 0.464                     & 0.420                     & 0.460                     & 0.380                          \\ 
\textit{[C+ob+op+att]} & pad.      & 0.490{$_{\pm0.002}$}          & 0.475$_{\pm0.052}$          & 0.174{$_{\pm0.010}$}          & 0.255{$_{\pm0.003}$}               \\ \bottomrule
\end{tabular}
\label{tb:crossvalidation-results}
\end{table}

\section{Conclusion}
\label{sec:conclusion}
In this paper, an end-to-end sequence-to-sequence generative model based on CVAE has been proposed to automatically detect interactions between vehicles and VRUs at intersections using video data. 
All the road users that appear during a vehicle's turning time are detected by a deep learning object detector, and their motion information is captured by optical flow, simultaneously. The sequences of object detection and optical--flow information together provide rich information for interaction detection. 
Both sliding window and padding methods are explored to learn dynamic patterns from turning sequences of varying lengths.
The proposed model predicts fine--grained interaction class label at each frame of less than \SI{0.1}{s}. which provides a clue of how the intensity of an interaction between a turning vehicle and VRUs evolves as time unfolds. 
The average voting scheme summarizes the frame--wise predictions so as to accurately get a class label for the overall sequence. 
Besides, the multi--sampling process generates diverse predictions and the Kernel Density Estimation function is used to measure the uncertainty level. 

The efficacy of the model was validated at a right--turn intersection in Germany and a left--turn intersection in Japan. It achieved an F1-score above 0.96 at the right--turn intersection and 0.89 at the left--turn intersection, and outperformed a sequence-to-sequence encoder--decoder model quantitatively and qualitatively. 

Furthermore, a series of ablation studies investigated the effectiveness of the combined information from object detection and optical flow, and the self-attention mechanism for learning temporal patterns from complex sequences. 
The comparison between the sliding window and padding methods showed that the former method is more flexible in coping with sequences of varying sequence length---the number of samples is not restricted to the maximum sequence length that a model can handle, which stands in contrast to the padding method. The self-attention mechanism has only shown a clear positive effect for interaction detection on the complex NGY dataset.

In future work, several improvements can be made to reduce the limitations of the detection model. First, the dichotomous classification of interaction should be extended to multi--class classification, \eg~taking the confrontation direction and car--following relationship into consideration. Second, the accuracy of feature extraction can be enhanced by using multiple cameras or even tracking. 
Third, projective transformation techniques or data recorded by drones with a bird's-eye view can be explored to reduce the distortion caused by the camera's perspective and filter the noisy optical--flow information captured from the through lane next to the turning lane. Last but not least, the generalizability of the model for interaction detection at different intersections needs to be further studied.

\section*{ACKNOWLEDGMENTS}
The project is funded by the German Research Foundation (DFG) through the Research Training Group SocialCars (227198829/GRK1931). This work is a collaboration with Murase Lab at Nagoya University and supported by Nagoya Toyopet Corporation with the Nagoya intersection dataset.

\bibliographystyle{IEEEtran}
\bibliography{bibliography}

\begin{IEEEbiography}[{\includegraphics[width=1in,height=1.25in,clip,keepaspectratio]{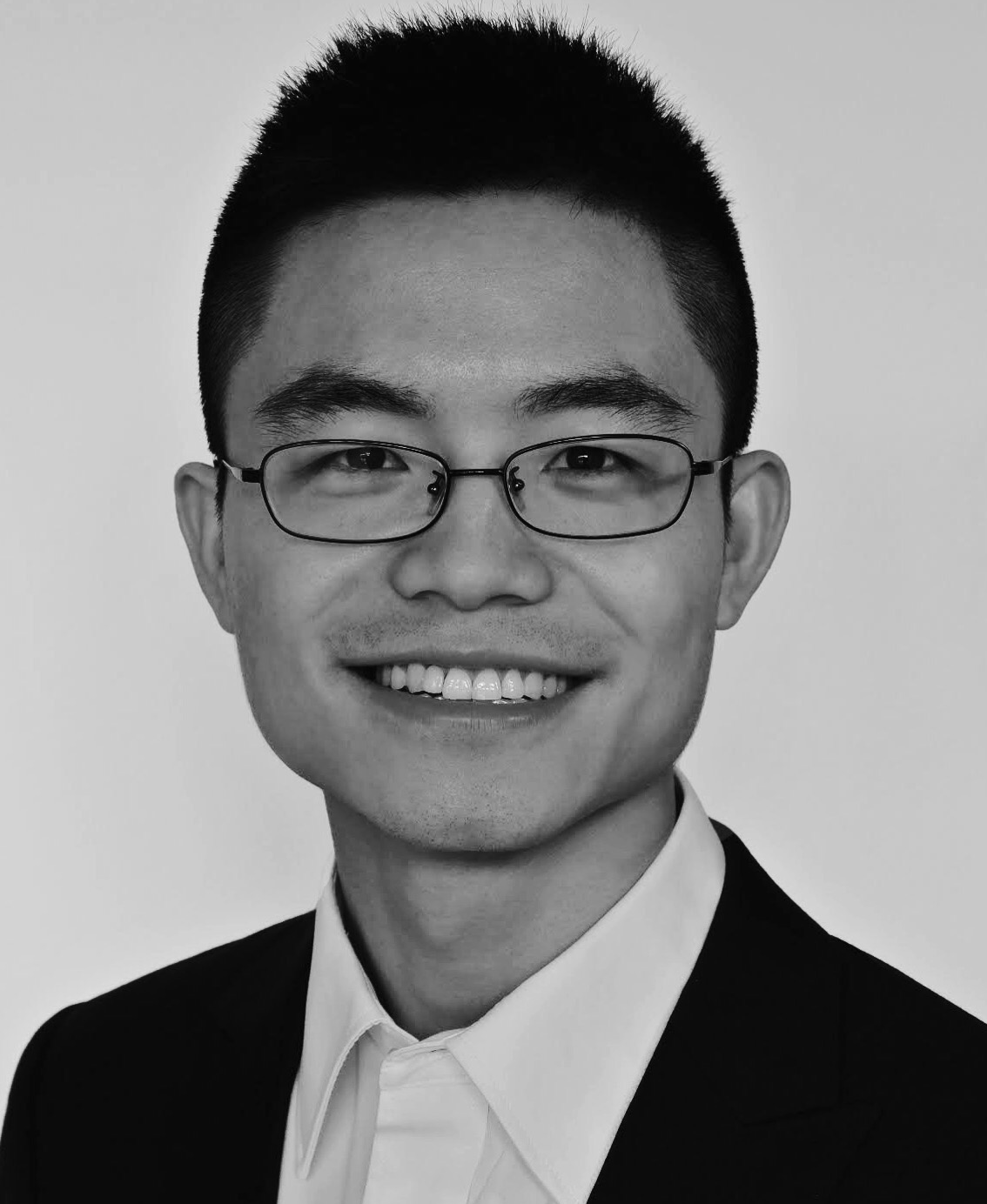}}]{Hao Cheng} received his B.A. degree in Information Management and Information System from Huazhong Agriculture University, China, in 2011, and his M.Sc. degree in Internet Technologies and Information Systems from Leibniz University Hannover, Germany, in 2017. He is currently finishing his Ph.D. at the Institute of Cartography and Geoinformatics at Leibniz University Hannover, Germany.

His research interests are deep learning and computer vision for road user behavior modeling in intelligent transport systems and autonomous driving.
\end{IEEEbiography}

\begin{IEEEbiography}[{\includegraphics[width=1in,height=1.25in,clip,keepaspectratio]{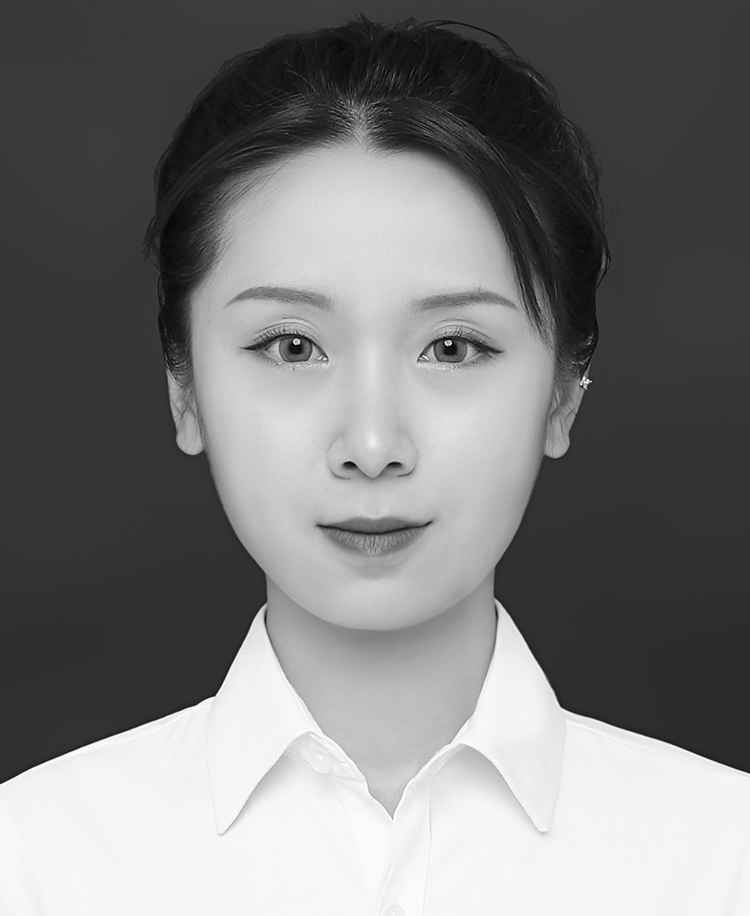}}]{Li Feng} received her B.Sc. Degree in Communication Engineering from Xidian University, China, in 2016, and her M.Sc. Degree in Electrical Engineering and Computer Science from Leibniz University Hannover, Germany, in 2020. 

She has been working as a deep learning algorithm engineer in China since graduation. Her main research interest is the visual task of lightweight object detection applied to Advanced Driver Assistance Systems.
\end{IEEEbiography}

\begin{IEEEbiography}
 [{\includegraphics[width=1in,height=1.25in,clip,keepaspectratio]{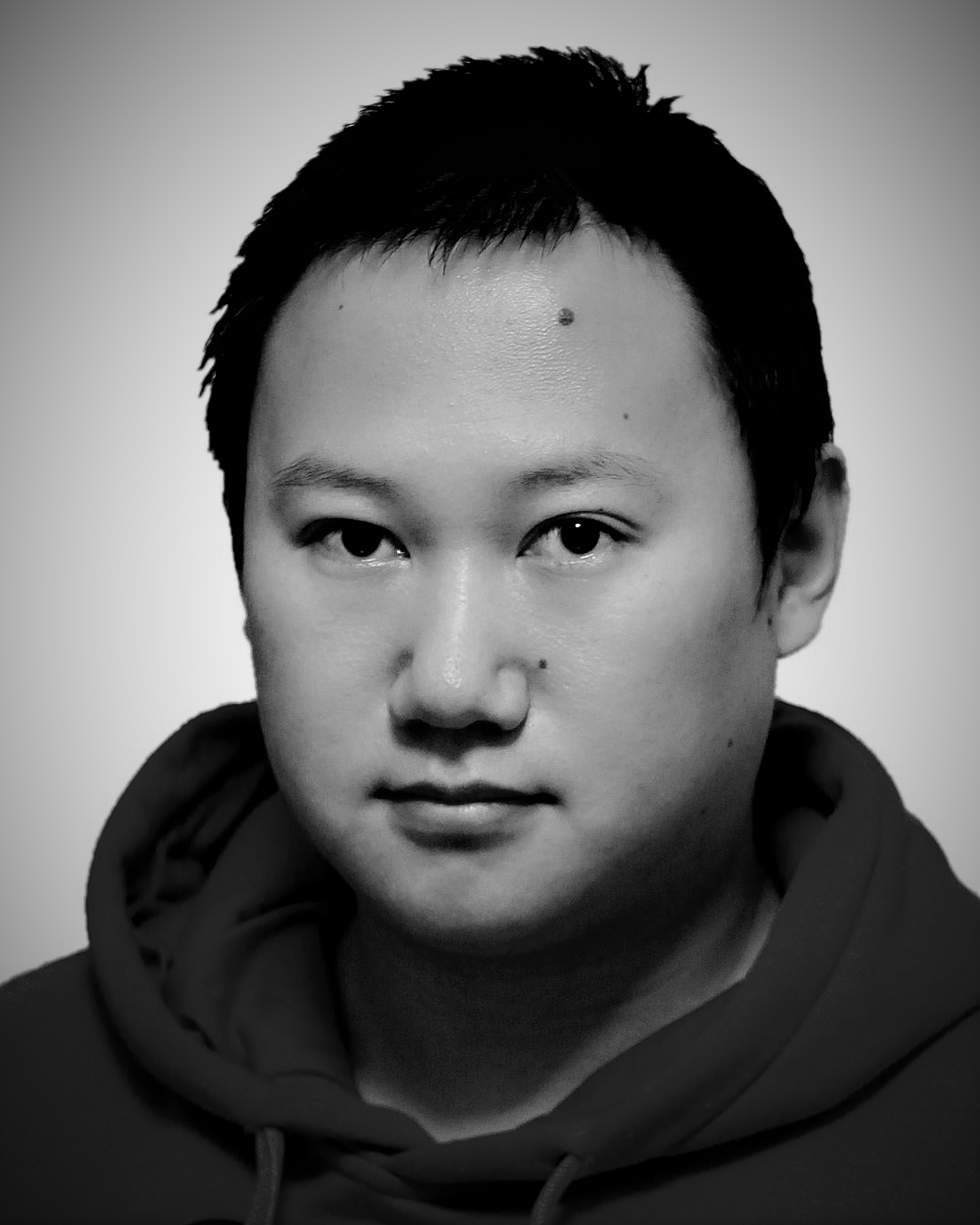}}]{Hailong~LIU} (Member, IEEE)
 received the M.Eng. and the Ph.D. degrees in Engineering from the Graduate School of Information Science and Engineering, Ritsumeikan University, Japan in 2015 and 2018, respectively. Meanwhile, he had been a Research Fellowship for Young Scientists (DC2) of JSPS from April 2016 to March 2018. From April 2018 to March 2019, he was a researcher of Institutes of Innovation for Future Society, Nagoya University, Japan. From April 2019, he is a researcher of the Graduate School of Informatics, Nagoya University, Japan. His research has focused on the representation learning for driving behaviors, human-machine interaction and solving the over-trust problem in the driving automation system. 
 He was awarded the IEEE IV (Intelligent Vehicles Symposium) Best Student Paper Award in 2015, and the IEEE GCCE (Global Conference on Consumer Electronics) Outstanding Paper Award in 2016.
 He is a member of IEEE, JSAE, JSAI and SICE.
\end{IEEEbiography}

\begin{IEEEbiography}
 [{\includegraphics[width=1in,height=1.25in,clip,keepaspectratio]{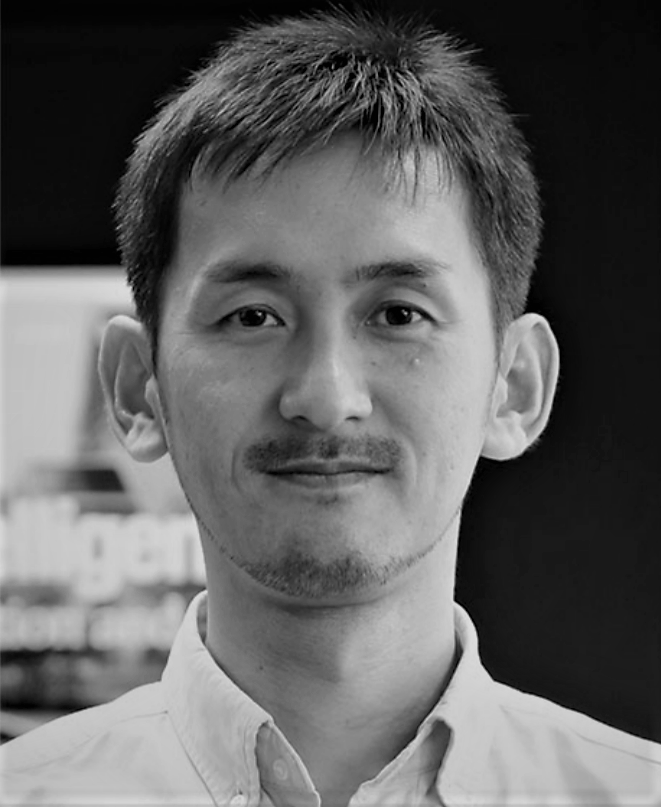}}]{Takatsugu~Hirayama} (Member, IEEE)
received the M.Eng. and Ph.D. degrees in Engineering Science from Osaka University in 2002 and 2005, respectively. From 2005 to 2011, he was a Research Assistant Professor at the Graduate
School of Informatics, Kyoto University. In 2011, he moved to the Graduate School of Information Science, Nagoya University as an Assistant Professor from 2012 to 2014, and a Designated Associate Professor from 2014 to 2017. He is currently a Designated Associate Professor at the Institutes of Innovation for Future Society, Nagoya University. His research interests include computer vision and human-computer interaction. He is a member of IEEE, ACM, IEICE, and IPSJ.
\end{IEEEbiography}

\begin{IEEEbiography}
 [{\includegraphics[width=1in,height=1.25in,clip,keepaspectratio]{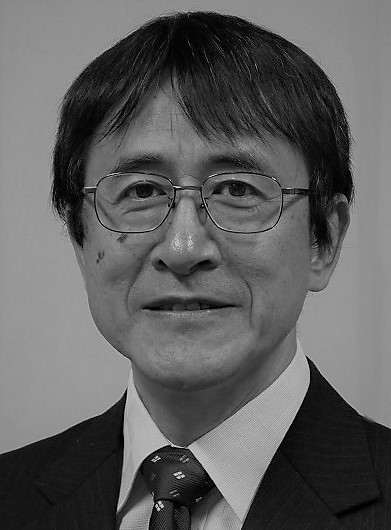}}]{Hiroshi~Murase} 
received his B.Eng., M.Eng., and Ph.D. degrees in Electrical Engineering from Nagoya University,
Japan. In 1980 he joined the Nippon Telegraph and Telephone Corporation (NTT). From 1992 to 1993, he was a visiting research scientist at Columbia University, New York. From 2003 he is a professor of Nagoya
University, Japan. He was awarded the IEICE Shinohara Award in 1986, the Telecom System Award in 1992,
the IEEE CVPR (Conference on Computer Vision and Pattern Recognition) Best Paper Award in 1994, the
IPS Japan Yamashita Award in 1995, the IEEE ICRA (International Conference on Robotics and Automation)
Best Video Award in 1996, the Takayanagi Memorial Award in 2001, the IEICE Achievement Award
in 2002, and the Ministry Award from the Ministry of Education, Culture, Sports, Science and Technology
in 2003. He is a Fellow of IEEE, IEICE, and IPS Japan.
\end{IEEEbiography}

\begin{IEEEbiography}[{\includegraphics[width=1in,height=1.25in,clip,keepaspectratio]{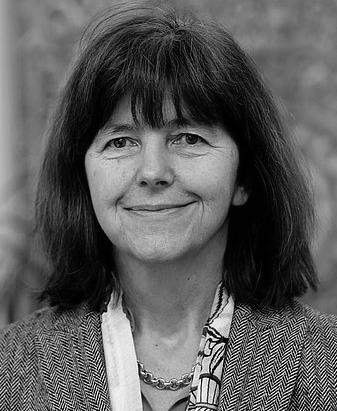}}]{Monika Sester} is a surveying engineer by training (University Karlsruhe) and earned her PhD on a topic of Machine Learning at the University of Stuttgart in 1995, and her habilitation in 2000 on the automatic generation of multiple representations of geodata. Since 2000 she has been a professor and head of the Institute of Cartography and Geoinformatics at Leibniz University Hannover. She has been Vice President of Leibniz University Hannover  (2016‐2017). Currently, she is the chair of a Senate Commission of the DFG on Earth System Science and member of the Senate of the Helmholtz Association.

She and her group work on the automation of spatial data processing with methods from computational, geometry, optimization and AI. Her projects are funded by the German Science Foundation (DFG), German Ministries, EU, as well as collaborations with Mapping Agencies and industry. She  chaired several Working Groups in ISPRS (International Society of Photogrammetry and Remote Sensing) and was  Vice  President  of  the  International  Cartographic  Association  (2015‐2019). 
\end{IEEEbiography}

\end{document}